\documentclass[acmsmall,natbib=false]{acmart}

\AtBeginDocument{%
  }

\setcopyright{acmlicensed}
\copyrightyear{2018}
\acmYear{2018}
\acmDOI{XXXXXXX.XXXXXXX}

\acmJournal{TWEB}
\acmVolume{37}
\acmNumber{4}
\acmArticle{111}
\acmMonth{8}

\RequirePackage[
  datamodel=acmdatamodel,
  style=acmauthoryear,
  ]{biblatex}
  
\usepackage{algpseudocode}
\usepackage[ruled,vlined,linesnumbered]{algorithm2e}
\usepackage{multirow}
\usepackage[caption=false]{subfig}

\begin{document}

\title{Full Triple Matcher: Integrating all triple elements between heterogeneous Knowledge Graphs}

\author{Victor Eiti Yamamoto}
\authornote{Both authors contributed equally to this research.}
\email{eitiyamamoto@nii.ac.jp}
\orcid{0000-0002-3825-6461}
\author{Hideaki Takeda}
\authornotemark[1]
\email{takeda@nii.ac.jp}
\orcid{0000-0002-2909-7163}
\affiliation{%
  \institution{National Institute of Informatics}
  \city{Chiyoda}
  \state{Tokyo}
  \country{Japan}
}
\affiliation{%
  \institution{Graduate University for Advanced Studies, SOKENDAI}
  \city{Hayama}
  \state{Kanagawa}
  \country{Japan}
}

\renewcommand{\shortauthors}{Yamamoto et al.}

\begin{abstract}
Knowledge graphs (KGs) are powerful tools for representing and reasoning over structured information. Their main components include schema, identity, and context. While schema and identity matching are well-established in ontology and entity matching research, context matching remains largely unexplored. This is particularly important because real-world KGs often vary significantly in source, size, and information density—factors not typically represented in the datasets on which current entity matching methods are evaluated. As a result, existing approaches may fall short in scenarios where diverse and complex contexts need to be integrated.

To address this gap, we propose a novel KG integration method consisting of label matching and triple matching. We use string manipulation, fuzzy matching, and vector similarity techniques to align entity and predicate labels. Next, we identify mappings between triples that convey comparable information, using these mappings to improve entity-matching accuracy. Our approach demonstrates competitive performance compared to leading systems in the OAEI competition and against supervised methods, achieving high accuracy across diverse test cases. Additionally, we introduce a new dataset derived from the benchmark dataset to evaluate the triple-matching step more comprehensively.
\end{abstract}



\received{20 February 2007}
\received[revised]{12 March 2009}
\received[accepted]{5 June 2009}

\maketitle

\section{Introduction} \label{section:introduction}

Knowledge graphs (KGs) use a graph-based data model to represent and manage knowledge from large-scale, heterogeneous data sources \cite{hogan2021knowledge}. Their structure provides a clear and flexible way to capture complex relationships between entities, making them valuable across a wide range of domains. General-purpose KGs such as DBpedia and YAGO cover diverse topics and entities, making them a useful starting point for retrieving information from different domains and supporting applications like question answering and information retrieval.

Large-scale KGs offer broad coverage but often lack the detailed and domain-specific information required for specialized applications such as recommendation systems \cite{noia2016}. To overcome this limitation, domain-specific KGs, which model high-level conceptualizations of particular domains such as healthcare or education, can be used to complement large-scale KGs. These domain-specific KGs are crucial for addressing problems that demand semantically rich, context-aware knowledge. However, leveraging both types of KGs effectively requires addressing the challenge of interoperability. As they are typically heterogeneous in structure and scope, integrating these KGs is essential to enable seamless knowledge sharing and application across domains.

The key components of a KG include schema, identity, and context \cite{hogan2021knowledge}. The schema provides a high-level structure for organizing data using tools such as RDF schemas, Shape Expressions, and ontologies. Identity refers to determining which nodes within the graph or across external sources represent the same real-world entity. Globally unique identifiers (e.g., IRIs) differentiate nodes to ensure clarity. At the same time, external identity links, such as the owl:sameAs property, indicate when a local entity corresponds to the same entity in another source. The context defines the conditions under which the data are valid. For example, certain information can be true at a specific time, point of view, or approximation. Even when the same data model and vocabulary are used, differences in context can emerge \cite{GuhaContext2004}. Reification, higher-arity representations, and annotations can capture context explicitly, but it is often implied rather than explicitly stated.

When integrating multiple KGs, it is essential to address key components: schema, identity, and context. Although schema and identity mapping are well studied, as seen in ontology matching \cite{shvaiko2005survey} and entity matching \cite{leone2022}, methods that effectively capture context remain underexplored. Context refers to the conditions under which data are valid — including implicit meaning derived from the assumptions, approximations and structured data \cite{GuhaContext2004}. Current approaches often emphasize structural and entity alignment, with limited attention to the subtleties of context. This research aims to identify similar and divergent triples between KGs to uncover shared and differing contexts, using these insights to enhance entity matching.

Various methods have been proposed to establish the mappings between entities in different KGs. Unsupervised and supervised methods have demonstrated excellent performance across multiple datasets. However, existing datasets lack real-world complexity, as they are often simplistic. For example, widely used data sets such as DBP15K exhibit similar scales and structures, with an overlapping ratio close to 100\%. However, real-world scenarios involve matching KGs from various sources, varying in size and information density. Moreover, matched entities typically represent only a tiny fraction of all entities \cite{jiang2023rethinking}. 

Matching heterogeneous KGs extracted from different sources presents additional challenges than matching KGs with similar origins. Heterogeneous KGs may differ substantially in their size, structure, and context. For example, KGs constructed from community-generated content can reflect different viewpoints and propositional attitudes toward the described domain. As a result, traditional methods relying solely on \textit{owl:sameAs} links and equivalent properties, such as \textit{owl:equivalentClass} and \textit{owl:equivalentProperty} for classes and properties, are insufficient for finding semantically similar triples. Furthermore, it is essential to ensure that the information across integrated KGs is compatible, not only at the entity and predicate level but also at the object level, requiring reconciliation of potential conflicts.

When two heterogeneous KGs are generated from different perspectives, it is rare for two entities to be identical. Therefore, we propose a heterogeneous matching approach emphasizing accepting compatible information rather than enforcing strict equivalence. We aim to match entities, entire triples, and graphs, as triple-level matching captures richer semantic relationships. Nonetheless, entity matching remains an integral part of the process.

To address this need, we introduce the triple matching problem, which aims to identify triples that convey semantically similar information. The relationship between matched triples can be classified as either compatible—where both triples can coexist without contradiction—or divergent—where the triples express conflicting information. Addressing the triple matching problem enables deeper integration of heterogeneous KGs by leveraging the semantic content of triples, moving beyond traditional entity or schema alignment.

In this research, we propose an algorithm to match all elements in a triple from heterogeneous KGs. The main objective is to receive two KGs and retrieve the triples representing compatible information. The algorithm has two main steps: label matching and triple matching. The first step creates mappings between entities and predicates from different KGs based on their labels. The second step uses the mappings obtained in the previous step to find triples that represent compatible information. The matched triple is also used to improve the matching between the entities. Our approach, called Full Triple Matcher (FTM), builds upon the PARIS matcher \cite{suchanek2011}, but we introduce modifications by adjusting its assumptions and incorporating the label matching and triple matching processes.


The similarity between two triples is computed by evaluating the triples' elements and the functionality of each predicate. A threshold is then applied to differentiate between triples with compatible and those with divergent information. We created a new dataset based on the KG track from the OAEI competition to evaluate FTM. We extracted all matching triples from the gold standard that involved predicates exhibiting a specific level of functionality. These triples were manually categorized as either conveying compatible or distinct content. FTM demonstrated high accuracy in identifying matching triples and accurately classifying them based on their content. We also evaluated our method for the entity matching task compared to supervised and unsupervised baselines, where we obtained competitive results to retrieve correct matchings. In our experiments, we used entity matching, as this task is more commonly addressed by existing baseline methods.

The contributions of this paper are as follows:
1- We proposed a method capable of obtaining results comparable to and outperforming those of state-of-the-art techniques and able to generate the mapping between triples.
2- We showed that using contextual information can improve the entity-matching process.
3- We propose a new task: A triple-matching problem.
4- We proposed a new dataset to evaluate the triple matching and achieved considerable results.

The remainder of this article is organized as follows: Section \ref{section:related-works} presents works related to this paper. Section \ref{section:definitions} defines terms used in this paper and the problem statement. Section \ref{section:model} explains this research's assumptions and proposed model. Section \ref{section:implementation} introduces our algorithm. Section \ref{section:experiment} shows the materials and obtained results. Section \ref{section:discussion} discusses the experiment and obtained results. Section \ref{section:conclusion} draws conclusions and future works.
\section{Related works} \label{section:related-works}

In this section, we review prior works relevant to our research. Subsection 2.1 discuss about schema matching related to KGs. Subsection \ref{subsect:entity_alignment} examines various entity matching methods, including those used for comparison with our approach, challenges addressed by different datasets, and alternative strategies to enhance entity alignment. Finally, Subsection \ref{subsect:triple_matching} explores methods for effectively resolving discrepancies between KGs and linking triples.

\subsection{Schema Matching} \label{subsect:schema_matching}

Schema matching is the process of identifying correspondences between different structural representations and plays a pivotal role in various domains, including description logic, databases, and graph data \cite{shvaiko2005survey}. Shvaiko et al. classify schema-matching techniques along two primary dimensions: granularity and input interpretation. These techniques can be further categorized based on the type of information they utilize: schema-level, instance-level, hybrid, and auxiliary approaches \cite{alwan2017survey}.

Schema-level approaches rely on intrinsic schema information, such as element names, data constraints, and structural relationships \cite{alwan2017survey}. Instance-level approaches, on the other hand, focus on analyzing data instances, making them particularly effective when schema metadata is limited or unreliable. Hybrid approaches integrate both schema- and instance-level information to enhance matching accuracy. Auxiliary approaches make use of external knowledge sources such as thesauri and dictionaries to support the matching process.

Notable examples of schema matching systems include CUPID \cite{madhavan2001generic} and AgreementMaker \cite{cruz2009agreementmaker}, both of which combine schema-level and auxiliary approaches by enriching schema feature information with thesauri. Conversely, PARIS \cite{suchanek2011} adopts an instance-level strategy, deriving predicate correspondences indirectly through instance matching, without relying on schema-level data.

In this work, we adopt a schema-level approach augmented with auxiliary information. Specifically, we utilize schema labels in combination with a pre-trained BERT model to enhance the semantic understanding of schema elements.

\subsection{Entity Alignment} \label{subsect:entity_alignment}

This subsection presents various methods for creating alignments between entities, including those we used as benchmarks for comparison in our experiments (Subsubsect. \ref{subsubsect:matching_methods}). Following that, we explore the processes and discussions surrounding the creation of different datasets (Subsubsect \ref{subsubsect:datasets}).

\subsubsection{Matching methods} \label{subsubsect:matching_methods}
Before the advent of neural approaches, statistical methods were widely used to address entity alignment problems \cite{leone2022}. These methods typically relied on data mining and database techniques to calculate the similarities of entities based on predefined rules and heuristics. Such methods are still explored in various fields, including the semantic web, where the OAEI competition \cite{hertling2020} is held annually, featuring both statistical and unsupervised techniques. We introduce the PARIS matcher \cite{suchanek2011}, which inspired our approach and demonstrates performance comparable to neural methods \cite{leone2022}. Additionally, we highlight LogMap \cite{jimenez2011logmap}, acknowledged as the top-performing matcher in the OAEI 2023 competition \cite{pour2023a}.

PARIS is an ontology matcher that uses a probabilistic approach to calculate the equivalence between predicates and entities \cite{suchanek2011}. The algorithm is based on the concept of functionality, which measures how likely a predicate can be interpreted as a function. Entity equivalence is determined through attribute values, which can be literals or other entities. For literals, exact matching is employed. When entities are the attribute values, their previously calculated equivalence probability is utilized. 
PARIS matcher inspired our approach to obtain the mapping between entities. However, the dependency on exact matches between literal values limits the use of KGs from different sources. Moreover, our testing shows that the original PARIS has high memory and time requirements to load and process large KGs. PARIS matcher is described with more detail in the subsection \ref{subsection:paris}.

LogMap is an ontology-matching tool that uses anchor mappings to start a mapping discovery and repairs steps using an ontology reasoner \cite{jimenez2011logmap}. The process begins by creating a lexical indexation based on the labels of the classes and enriching it using an external lexicon. Second, the algorithm creates a class hierarchy for each input KG and uses a reasoner to extend the hierarchy with disjoint information. Third, the algorithm uses the lexical index to generate the mapping between the entities and calculate a confidence score based on the matching neighbors with similar strings. Fourth, new mappings are discovered, expanding the context of anchor mapping and calculating the string similarity, and repaired, using Horn clauses to calculate if the mapping is satisfiable. LogMap obtains a high-precision mapping between entities. Still, it depends on a high-quality class ontology to work, making it unsuitable for community-based KG that lacks a structured ontology.

Neural methods represent a newer approach to the entity matching problem, sparking renewed interest in the field \cite{leone2022}. The matching process typically involves four key steps: embedding, relation, attribute, and alignment modules \cite{huang2024representation}. The embedding module maps elements of a knowledge graph (KG) into a vector space, primarily using translational models or graph convolutional network (GCN)-based models. Translational models treat relationships as transformations from the head entity to the tail entity, while GCN-based models aggregate information from neighboring nodes using graph convolutional techniques. The relation and attribute modules enhance the embeddings by leveraging entities’ relationships and attributes. Finally, the alignment module computes the similarity between entity pairs based on the embeddings. We introduce IMUSE as one of the first neural approaches to integrate attributes and relationships for entity matching. Additionally, we highlight AttrE and NMN as the most effective matchers for translational and GCN-based models, respectively \cite{zhang2022benchmark}. Lastly, we present AutoAlign as an example of an unsupervised method for entity matching.

IMUSE generates initial entity pairs as seeds based on attributes and embeds entities using relationships. String attributes utilize Levenshtein distance and the longest common subsequence problem to compute similarity, while numeric attributes are normalized for comparison. Predicate pairs are generated by matching objects. The relationships are embedded using a translational model, and a regression model is applied to weigh the similarity derived from both attributes and relationships. This approach was tested on a subset of KGs created using Baidu, Wikipedia, and the Hudon Encyclopedia, focusing on entities related to people, places, and living things. While the method produced promising results, other approaches outperformed it in a more recent study \cite{OpenEA}.

AttrE is a matcher designed to align entities in KGs by leveraging their attributes \cite{trisedya2019entity}. Unlike traditional methods primarily focusing on relationships between entities, AttrE incorporates attribute information, which is often a rich and underutilized resource in KGs. This technique involves three main steps: predicate alignment, embedding learning, and entity alignment. In predicate alignment, the method standardizes the names of predicates from two KGs to place them into a common vector space. This step helps represent the relationships consistently across different KGs. The embedding learning process combines the unified relationship vector space with attribute character embeddings to form a unified entity vector space. The approach is inspired by the TransE method \cite{bordes2013translating}, but predicates are seen as translations from one entity to an attribute. AttrE uses a combination of character embeddings, LSTM-based functions, and N-gram-based functions to embed attributes. Finally, entity alignment utilizes the previous steps' structure and attribute embeddings to map similar entities from different KGs in the unified entity vector space. This mapping helps align entities with similar characteristics or meanings across KGs. AttrE obtained outstanding results compared to other translational models by leveraging attribute information \cite{zhang2022benchmark}. However, the accuracy degrades when the number of entities in the dataset increases.

The Neighborhood Matching Network (NMN) is an entity alignment technique based on graph convolutional networks (GCNs) \cite{wu-2020-neighborhood}. GCN-based methods operate under the assumption that entities with similar characteristics also have similar neighboring structures. However, real-world scenarios often involve entities representing the same concept yet having different neighbor sizes and topological structures due to the incomplete and heterogeneous nature of KGs. The NMN addresses this challenge by capturing the most relevant neighbors to estimate similarities between entities across different KGs. This approach involves four main steps: KG structure embedding, neighborhood sampling, neighborhood matching, and neighborhood aggregation. In the KG structure embedding step, the technique learns the KG's structural embedding using multi-layered GCNs and initializes with pre-trained word embeddings. The neighborhood sampling step aims to select the most informative entities by prioritizing neighbors contextually more relevant to the central entity. The neighborhood matching step seeks to identify which neighbors of the analyzed entity closely relate to neighbor entities in the target KGs. The method employs a cross-graph matching vector to embed each neighbor to achieve this. Finally, the neighborhood aggregation step combines information from the KG structure embedding and the neighborhood matching stage to generate embeddings used for alignment. NMN obtained the best result among GCN methods for the DWY-NB dataset \cite{zhang2022benchmark}. Still, an extensive training set is required to get a comparable result to translational methods.


AutoAlign is an unsupervised KG matcher that aligns predicates by constructing a predicate-proximity graph using a large language model (LLM). It learns the KG representation independently and then unifies them into the same vector space using attributes as anchors \cite{zhang2023autoalign}. The algorithm operates in three stages: predicate embedding, attribute embedding, and structure embedding. In the predicate embedding stage, the subjects and objects of triples are replaced by their types using <rdfs:type>, and a large language model standardizes predicate names. For structure embedding, TransE is employed with weights influenced by the predicate mappings from the predicate embedding stage. In the attribute embedding stage, attribute values are transformed into vectors using a sum compositional function, LSTM-based function, or n-gram-based function. While predicate embedding unifies the vector space, entity embedding is done separately. Attribute embedding is then used to map entity embeddings into the same vector space.

Recent trends in entity alignment have increasingly incorporated multimodal KGs and large language models (LLMs). Methods such as MDSEA \cite{fang2024mdsea} and MSEEA \cite{wang2025multi} extend beyond purely textual KGs, integrating additional modalities such as images and videos. MDSEA learns entity representations by leveraging entity attributes, relations, graph structure, and associated images \cite{fang2024mdsea}. Similarly, MSEEA generates embeddings for entities while also enriching them with text descriptions derived from images and transforming triples into descriptive sentences \cite{wang2025multi}. As multimodal integration lies beyond the scope of our current work, we did not include comparisons with these approaches.

LLMs are also being increasingly applied to the entity alignment task. For instance, AutoAlign \cite{zhang2023autoalign} utilizes an LLM to normalize predicate names. LLMEA \cite{yang2024two} employs LLMs to translate entity labels into English and formulates multiple-choice questions to guide entity-matching decisions. LLM-Align \cite{chen2024llm} identifies the most informative attributes and relations, which are then fed into the LLM to facilitate alignment. Since our experiments do not involve cross-lingual datasets, we chose to compare our results specifically with AutoAlign.

Other studies have also sought to enhance supervised methods. For example, DAAKG utilizes active learning to generate batches for human annotation \cite{Huang2023DAAKG}, while Guo aims to discover neural axioms that inherently connect specific classes to predicates within the vector space learned during the embedding process \cite{pmlr-v162-guo22i}. However, both approaches lie beyond the scope of this study.

Neural approaches have been the focus of newer research, but they have high time and memory requirements, making them impractical for real-world applications \cite{leone2022}. Additionally, supervised techniques require a large amount of training data, which may not always be available. Statistical methods like PARIS can achieve similar or even better results than neural approaches. However, they are typically based on lexical similarity, which makes them unsuitable for matching cross-lingual or domain-specific knowledge graphs. Additionally, PARIS performs poorly on datasets lacking attribute triples, which can be expected for entities with incomplete information. 
We used AttrE, NMN, LogMap, PARIS, and AutoAlign during our experiments to compare the performance with our approach (Section \ref{section:experiment}).

\subsubsection{Datasets} \label{subsubsect:datasets}
Various datasets have been proposed to evaluate the effectiveness of different entity-matching methods. However, a dataset represents an approximation of real-world scenarios, with varying degrees of success in doing so. Additionally, datasets can be modified to assess the strengths and weaknesses of entity-matching approaches.

Huang et al. observed that supervised entity alignment algorithms typically comprise multiple modules, but their evaluation focuses solely on the overall process \cite{huang2024representation}. They further noted that while the datasets used for testing are often information-dense, real-world cases can be noisy and incomplete. Their work identified a framework for entity alignment algorithms comprising four main modules: an embedding module, a relation module, an attribute module, and an alignment module. The embedding module initiates the process by converting the knowledge graph into vectors. The relation module then structures the vector space based on entity relationships, while the attribute module incorporates entity attributes into the vector space. Finally, the alignment module calculates the similarity between entities based on the embeddings. They also developed a set of datasets to assess various interactions between the modules and to tackle more complex scenarios. Their findings suggest that simple label matching can suffice in certain cases, while the effectiveness of different strategies depends on the dataset's characteristics. Since our method is not supervised, it does not follow the framework outlined in their study. However, we evaluated our entity alignment algorithm using the entire DBpedia dataset to assess its performance in real-world scenarios. We also compared the performance with a baseline matcher that only uses label matching to check whether the KG matches can beat the simplest label-matching cases.

Jiang et al. argue that current datasets are overly simplistic and lack the complexity found in real-world applications \cite{jiang2023rethinking}. Widely used datasets, such as DBP15K, share similar scales and structures, with overlapping ratios nearing 100\%. GNN-based methods attempt to leverage structural information to align entities with similar structures, achieving high performance on these datasets. However, in practical scenarios, KGs from different sources often vary significantly in size, density, and structure. Furthermore, matched entities typically represent only a small subset of the entire KG. To address these challenges, the authors introduced a new dataset combining domain-specific KG (ICEWS) with general KGs (WIKIDATA, YAGO). They proposed a matching method incorporating entity names and temporal encoders. Their evaluation, which included GNN-based methods and models like BERT-INT, revealed that existing approaches struggle to effectively integrate structural and semantic information. GNN-based methods performed poorly when faced with differing KG structures, while methods like BERT-INT primarily relied on entity names. Although we did not use their proposed dataset due to its reliance on triples annotated with temporal data, which is beyond the scope of our work, we evaluated our method on datasets derived from various sources and construction techniques. This allowed us to demonstrate its effectiveness across different scales and structural complexities.

Zhang et al. highlight several limitations in current datasets, including the assumption of bijection, small scale, and limited name diversity \cite{zhang2022benchmark}. To address these issues, they modified the datasets by removing a certain percentage of corresponding entities, altering entity names using alternative attribute values, and generating a larger dataset version. Their experiments demonstrated that matching approaches based on attributes and relationships significantly improve performance. However, they also identified scalability challenges when applied to larger datasets.

In our research, we developed a statistical entity matching method that operates without requiring training seeds and demonstrates strong performance on real-world KGs of significant size, surpassing existing methods. To evaluate our approach, we utilized the dataset from \cite{zhang2022benchmark}, comparing it against supervised methods in scenarios where the cases were non-bijective, label-dependent, or small in scale. Additionally, we tested our method on datasets from the OAEI competition \cite{hertling2020} and a subset of \cite{hertling2022gollum}, leveraging KGs with diverse sources and structures to ensure robustness.

\subsection{Triple matching} \label{subsect:triple_matching}

The triple-matching task has not been directly explored as a primary research focus. However, various approaches attempt to identify similar triples to achieve other objectives, such as detecting misinformation or completing a KG. Additionally, calculating the similarity between different literal types and values remains an open challenge.

Koudas et al. discuss using a general functional dependency to merge databases from heterogeneous sources \cite{koudas2009metric}. The idea is to create a more robust method for differences caused by changes in representation conventions, such as how movie durations are measured or abbreviations in the addresses. The proposed method uses Euclidean distances for numbers and cosine similarity on high-dimensional q-gram vectors for strings. The technique was tested on movie duration, latitude, longitude, and string integration. In our approach, we applied our method to the KG context and tested different predicates.

Liu et al. presented an algorithm to identify misinformation in linked KGs by evaluating the accuracy of triples \cite{Liu2017}. Their method begins by retrieving links equivalent to the subject of the triple using the "sameAs" service. These links are then used to extract predicates and objects from the linked KGs through SPARQL queries. Next, the algorithm exploits WordNet to match triples based on the least common subsumer (LCS) between concepts, filtering out mismatched predicates through predicate type comparison. Finally, a confidence score is computed by comparing the retrieved information to the weighted average of property values from matched triples. While the proposed technique demonstrates a high f-measure, its dataset was restricted to birth and death dates. In contrast, our approach offers the additional benefit of uncovering previously unknown entity links.

Munne and Ichise proposed a methodology for completing KG, employing a secondary KG to establish a ranking system for crucial attributes and missing values intended for incorporation into the target KG \cite{munne2023}. The fundamental concept involves selecting the most significant characteristics from the source KG and suggesting them to analogous entities within the target KG. This process utilizes embedding-based ranking, popular-based ranking, and stochastic gradient descent to optimize both rankings. They employed a gated recurrent unit to train a model to identify similar attributes across KGs. Subsequently, they aligned triples to eliminate redundant information in the target KG. The primary goal of this research was to complement the target KG, yet it also facilitated the identification of inconsistencies between KGs by comparing triples. The authors tested their approach using DBpedia and Yago, revealing inconsistencies of 9.45\% in death dates, 6.70\% in founding dates, and 5.02\% in birth dates. The proposed method shows that even KGs from similar sources can have inconsistencies. Our approach focuses on finding consistent and inconsistent information between the mapped KGs, and our method can compare different data types.

In this study, we propose a novel task and method for aligning triples in order to identify semantically similar or conflicting information. To the best of our knowledge, no existing method directly addresses this task, and therefore, no comparable baseline is available. Additionally, we introduce a new dataset that extends the variety of triple types used in previous works by incorporating diverse combinations of data types, including both attribute and relationship triples.
\section{Definitions} \label{section:definitions}
This section defines the terms used in this paper and the problem statement.

\noindent
\textbf{Knowledge Graph.} We denote knowledge graphs as $\mathcal{G} =(\mathcal{E}, \mathcal{P}, \mathcal{L}, \mathcal{T})$ where $e \in \mathcal{E}$ is an entity, $p \in \mathcal{P}$ is a predicate, $l \in \mathcal{L}$ is a literal, and $t = (s, p, o) \in \mathcal{T}$ is a triple from $\mathcal{T} \subset \mathcal{E} \times \mathcal{P} \times (\mathcal{E} \cup \mathcal{L})$, where $s$ is the subject, $p$ is the predicate and $o$ is the object.

\noindent
\textbf{Entity alignment problem.} Given two KGs $\mathcal{G}_1 = (\mathcal{E}_1, \mathcal{P}_1, \mathcal{L}_1, \mathcal{T}_1)$ and $\mathcal{G}_2 = (\mathcal{E}_2, \mathcal{P}_2, \mathcal{L}_2, \mathcal{T}_2)$, the objective is to create a set of mappings $m_e \in \mathcal{M}_e$ where $m_e = (e_1, e_2, c)$, entities $e1 \in \mathcal{E}_1$ and $e_2 \in \mathcal{E}_2$, and $c \in \mathbb{R}$ represents the confidence of the equivalence between $e_1$ and $e_2$.

\noindent
\textbf{Predicate alignment problem.} Given two KGs $\mathcal{G}_1 = (\mathcal{E}_1, \mathcal{P}_1, \mathcal{L}_1, \mathcal{T}_1)$ and $\mathcal{G}_2 = (\mathcal{E}_2, \mathcal{P}_2, \mathcal{L}_2, \mathcal{T}_2)$, the objective is to create a set of mappings $m_p \in \mathcal{M}_p$ where $m_p = (p_1, p_2, c)$, predicates $p1 \in \mathcal{P}_1$ and $p_2 \in \mathcal{P}_2$, and $c \in \mathbb{R}$ represents the confidence of the equivalence between $p_1$ and $p_2$.

\noindent
\textbf{Object similarity problem.} Given two KGs $\mathcal{G}_1 = (\mathcal{E}_1, \mathcal{P}_1, \mathcal{L}_1, \mathcal{T}_1)$ and $\mathcal{G}_2 = (\mathcal{E}_2, \mathcal{P}_2, \mathcal{L}_2, \mathcal{T}_2)$, the objective is to calculate the similarity $s \in \mathcal{R}$ of two objects $o_1 \in \mathcal{E}_1 \cup \mathcal{L}_1$ and $o_2 \in \mathcal{E}_2 \cup \mathcal{L}_2$.

\noindent
\textbf{Triple matching.} Given two KGs $\mathcal{G}_1 = (\mathcal{E}_1, \mathcal{P}_1, \mathcal{L}_1, \mathcal{T}_1)$ and $\mathcal{G}_2 = (\mathcal{E}_2, \mathcal{P}_2, \mathcal{L}_2, \mathcal{T}_2)$, the objective is to create a set of triples mappings $m_t \in \mathcal{M}_t$ where $m_t = (t_1, t_2, r)$, $t_1 \in \mathcal{T}_1$, $t_2 \in \mathcal{T}_2$, and $r$ is the relationship between the triples. The relationship can be compatible or divergent.

\noindent
\textbf{Functionality.} Suchanek et al. proposed the concept of functionality to represent how likely a predicate can be interpreted as a function in a KG $\mathcal{G}$ \cite{suchanek2011}. Given $p(x, y)$ as a representation of a triple $(x, p, y) \in \mathcal{T}$ and $\#y : \delta(y)$ as a representation of $|\{y | \delta(y)\}|$. 

{
\begin{equation}
    fun(p) = \frac{\#x: \exists y. p(x, y)}{\#x,y : p(x, y)}
    \label{eq:functionality}
\end{equation}
}

An inverse functionality was also proposed, which is defined as $fun^{-1}(p) = fun(p^{-1})$ \cite{suchanek2011}.

\noindent
\textbf{Unique ratio.} The unique ratio represents the variety of object values associated with a predicate in the KG. We define unique ratio as the size of unique object values divided by the size of triples related to the analyzed predicate (Equation \ref{eq:unique_ratio}).

{
\begin{equation}
    uniqueRatio(p) = \frac{|set(t_o)|}{|t|} | t = (t_s, t_p, t_o), t_p = p
    \label{eq:unique_ratio}
\end{equation}
}
\section{Proposed Model} \label{section:model}
This section introduces the model employed by the PARIS algorithm, which significantly influenced our research, as outlined in Subsection \ref{subsection:paris}. Subsequently, we present FTM’s assumptions and proposed model in Subsection \ref{subsect:full_triple_matcher_model}. Finally, we outline the difference between PARIS model and Full Triple Matcher in Subsection \ref{subsect:comparison}

\subsection{PARIS Model} \label{subsection:paris}
PARIS is an ontology-matching algorithm that can also match KGs \cite{suchanek2011}. The algorithm receives as input two KGs $\mathcal{G}_1 = (\mathcal{E}_1, \mathcal{P}_1, \mathcal{L}_1, \mathcal{T}_1)$ and $\mathcal{G}_2 = (\mathcal{E}_2, \mathcal{P}_2, \mathcal{L}_2, \mathcal{T}_2)$ and returns as output a set of mappings $m_p \in \mathcal{M_P}$. Each mapping $m$ is a triple $(e_1, e_2, c)$ where $e_1 \in \mathcal{E}_1$, $e_2 \in \mathcal{E}_2$ and $c \in \mathbb{R}$ representing the confidence of their equivalence.



Each KG has a different set of predicates. It is possible to calculate the probability $Pr(p_1 \subset p_2)$ of predicate $p_1$ being a sub-relation of $p_2$ using the equation \ref{eq:subr-1}.

{
\begin{equation}
\label{eq:subr-1}
    Pr(p_1 \subset p_2) = \frac{\sum_{p_1(e_1, y_1)} (1- \prod_{p_2(e_2, y_2)} (1-(Pr(e_1 \equiv e_2) \times Pr(y_1 \equiv y_2))))}{\sum_{e_1, y_1} (1-\prod_{e_2, y_2} (1-(Pr(e_1 \equiv e_2) \times Pr(y_1 \equiv y_2))))}
\end{equation}
}

\normalsize
PARIS proposes that two entities $e_1$ and $e_2$ have high equivalence probability if triples $(e_1, p, y_1) \in \mathcal{T}_1$ and $(e_2, p, y_2) \in \mathcal{T}_2$ exist, one of the predicates is a sub-relation of another one, predicate $p_1$ or $p_2$ is highly inverse functional and if $y_1 \equiv y_2$. This idea can be represented as the equation \ref{eq:prob-final}.

{
\begin{multline}
\label{eq:prob-final}
        Pr(e_1 \equiv e_2) = 1 - \prod_{p(e_1, y_1)} \prod_{p(e_2, y_2)} (1 - Pr(p_1 \subset p_2) \times fun^{-1}(p_1) \times Pr(y_1 \equiv y_2)) \times \\
 (1 - Pr(p_2 \subset p_1) \times fun^{-1}(p_2) \times Pr(y_1 \equiv y_2)) \\
\end{multline}
}

\normalsize
Note that equations \ref{eq:subr-1} and \ref{eq:prob-final} have a recursive dependency. The algorithm iterates the calculation of probability between entities and probability between predicates. In the first iteration, all pairs of predicates start with the same probability $\theta = 0.1$, and it is iterated $n$ times.

The elements $y_1$ and $y_2$ can be entities or literals. The algorithm uses the probability between entities to calculate $Pr(y_1 \equiv y_2)$ if they are entities. The algorithm uses an exact match between them if they are literals. It means that $Pr(y_1 \equiv y_2) = 1$ if $y_1$ and $y_2$ are equivalents and $0$ otherwise.

\subsection{Full Triple Matcher (FTM) Model} \label{subsect:full_triple_matcher_model}

\begin{figure}
    \centering
    \includegraphics[width=\columnwidth]{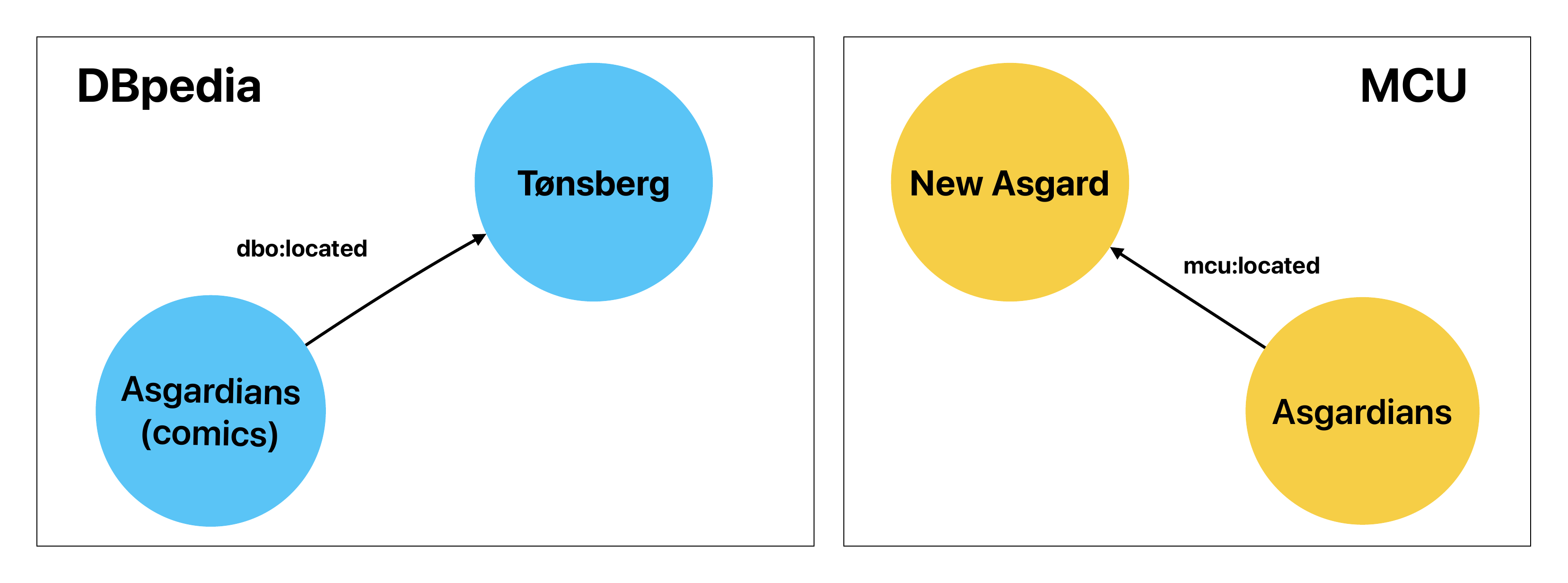}
    \Description[High functionality example]{High functionality example}
    \caption{Triple example with high functionality predicate}
    \label{fig:high-functionality}
\end{figure}

\begin{figure}
    \centering
    \includegraphics[width=\columnwidth]{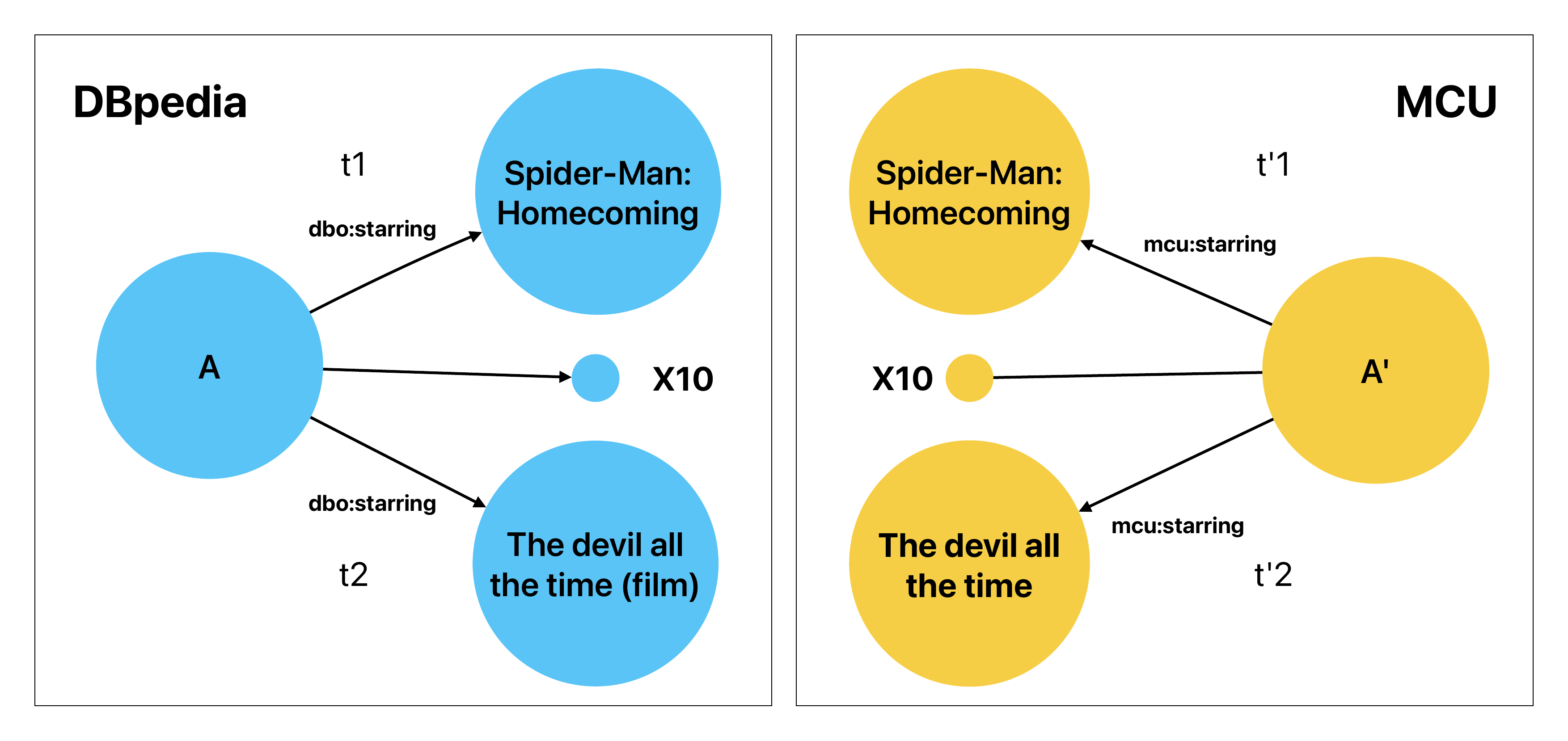}
    \Description[Low functionality example]{Low functionality example}
    \caption{Triple example with low functionality predicate}
    \label{fig:low-functionality}
\end{figure}

We assume that two triples can be considered similar if their subjects, predicates, and objects demonstrate a high similarity and both predicates exhibit high functionality or inverse functionality. We believe that predicates with high functionality (n-1) or inverse functionality (1-n) can better represent the characteristics of an entity. For instance, predicates such as "a person is a citizen of a country" (n-1) or "a person is the mother of another person" (1-n) are given more weight compared to "a person acted in a movie" (n-n). Nevertheless, we also account for n-n triples, as although each triple carries less weight, a set of similar n-n triples can collectively indicate a meaningful similarity between two entities. Figures \ref{fig:high-functionality} and \ref{fig:low-functionality} represent cases where the triples have high and low functionality, respectively.

Figure \ref{fig:high-functionality} illustrates a case involving a triple with a predicate that exhibits near n-1 cardinality. This occurs because an entity is typically associated with a single location. As a result, the predicate shows high functionality (0.79 for \textit{dbo:located} and 0.84 for \textit{mcu:located}) and low inverse functionality (0.16 for \textit{dbo:located} and 0.20 for \textit{mcu:located}). The underlying assumption is that if two subjects refer to the same or closely related entities, and both are associated with a particular location, then the corresponding objects may also represent similar concepts. In terms of triples, if two triples convey a similar idea and the subject entities (e.g., \textit{Asgardians (comics)} and \textit{Asgardians}) are semantically close, then the object entities (e.g., \textit{Tønsberg} and \textit{New Asgard}) may also refer to related notions.

In contrast, Figure \ref{fig:low-functionality} presents a scenario with a relationship that approaches n-n cardinality. For instance, an actor may appear in multiple films, and each film may feature multiple actors. In this case, the predicates exhibit both low functionality and low inverse functionality. Consequently, the similarity between such triples tends to be lower, as one cannot reliably conclude that two actors represent the same person merely because they appear in the same film, nor that two films are related simply because they share the same actor.

Equations \ref{eq:fun_triples} and \ref{eq:invfun_triples} present the probability of two triples being similar when their predicates are highly functional or inverse functional, respectively. The final similarity between the two triples is calculated using Equation \ref{eq:triple_matching}, which assesses whether Equations \ref{eq:fun_triples} or \ref{eq:invfun_triples} show high similarity.

The mapping of triples forms a set of triple pairs, and the probability is determined using Equation \ref{eq:triple_matching}. In formal terms, this means that each pair of triples and their similarity probability, $(t_1, t_2, Pr^{triple}(t_1, t_2))$, belongs to the mapping set $M_{t}$.

\begin{equation}
    Pr^{triple}_{fun}(t_1, t_2) = Pr^{ent.}(s_1, s_2) \times Pr^{pred.}(p_1, p_2) \times fun(p_1) \times fun(p_2) \times Pr^{obj.}(o_1, o_2)
    \label{eq:fun_triples}
\end{equation}

\begin{equation}
    Pr^{triple}_{invFun}(t_1, t_2) = Pr^{ent.}(s_1, s_2) \times Pr^{pred.}(p_1, p_2) \times invFun(p_1) \times invFun(p_2) \times Pr^{obj.}(o_1, o_2)
    \label{eq:invfun_triples}
\end{equation}

\begin{equation}
    Pr^{triple}(t_1, t_2) = 1 - (1 - Pr^{triple}_{fun}(t_1, t_2)) \times (1 - Pr^{triple}_{invFun}(t_1, t_2))
    \label{eq:triple_matching}
\end{equation}

Consider the example illustrated in Figure \ref{fig:high-functionality}. Assume that the similarity between the predicates \textit{dbo:located} and \textit{mcu:located} has already been computed as 1.0. Additionally, the functionality and inverse functionality values for these predicates are as follows: \textit{dbo:located} has a functionality of 0.79 and an inverse functionality of 0.16, while \textit{mcu:located} has a functionality of 0.84 and an inverse functionality of 0.20.

The entity similarity between \textit{Asgardians (comics)} and \textit{Asgardians} is 0.9, and between \textit{Tønsberg} and \textit{New Asgard} is 0.5. Our goal is to compute the similarity between the triples \textit{(\textit{Asgardians (comics)}, \textit{dbo:located}, \textit{Tønsberg})} and \textit{(\textit{Asgardians}, \textit{mcu:located}, \textit{New Asgard})}.

Using Equations~\ref{eq:fun_triples} and~\ref{eq:invfun_triples}, we obtain similarity scores of 0.16 (Equation~\ref{eq:high_fun_div_triples}) and 0.01 (Equation~\ref{eq:high_inv_fun_div_triples}), respectively. By combining the functionality and inverse functionality similarities, the resulting triple similarity is 0.30, as computed using Equation~\ref{eq:high_triple_matching}.

\begin{equation}
    Pr^{triple}_{fun}(t_1, t_2) = 0.9 \times 1.0 \times 0.79 \times 0.84 \times 0.5 = 0.29
    \label{eq:high_fun_div_triples}
\end{equation}

\begin{equation}
    Pr^{triple}_{invFun}(t_1, t_2) = 0.9 \times 1.0 \times 0.16 \times 0.20 \times 0.5 = 0.01
    \label{eq:high_inv_fun_div_triples}
\end{equation}

\begin{equation}
    Pr^{triple}(t_1, t_2) = 1 - (1 - 0.29) \times (1 - 0.01) = 1 - 0.71 \times 0.99 = 0.30
    \label{eq:high_triple_matching}
\end{equation}

Using the same idea, we can calculate the probability of two triples having divergent information. To represent this idea, we define that two triples are divergent if their subject and predicate are highly similar; both predicates exhibit high functionality or inverse functionality, but their objects have low similarity. Equations \ref{eq:fun_div_triples} and \ref{eq:invfun_div_triples} calculate the divergence for high functionality and high inverse functionality, respectively. Equation \ref{eq:triple_div_matching} represents the final equation to obtain the probability of two triples being divergent.

\begin{equation}
    Pr^{divTriple}_{fun}(t_1, t_2) = Pr^{ent.}(s_1, s_2) \times Pr^{pred.}(p_1, p_2) \times fun(p_1) \times fun(p_2) \times (1-Pr^{obj.}(o_1, o_2))
    \label{eq:fun_div_triples}
\end{equation}

\begin{equation}
    Pr^{divTriple}_{invFun}(t_1, t_2) = Pr^{ent.}(s_1, s_2) \times Pr^{pred.}(p_1, p_2) \times invFun(p_1) \times invFun(p_2) \times (1-Pr^{obj.}(o_1, o_2))
    \label{eq:invfun_div_triples}
\end{equation}

\begin{equation}
    Pr^{divTriple}(t_1, t_2) = 1 - (1 - Pr^{divTriple}_{fun}(t_1, t_2)) \times (1 - Pr^{divTriple}_{invFun}(t_1, t_2))
    \label{eq:triple_div_matching}
\end{equation}

We consider that two entities are similar if they have a matching triple where the subject or the object is the matching pair. It means that a pair of entities $e_1$ and $e_2$ are similar entities if exists a triple mapping $(t_1, t_2, Pr^{triple}(t_1, t_2)) \in M_{t(e_1, e_2)} = \{(t_1 = (e_1, p_1, o_1) \wedge t_2 = (e_2, p_2, o_2)) \vee (t_1 = (s_1, p_1, e_1) \wedge t_2 = (s_2, p_2, e_2))\}$. The probability of two similar entities based on the triples is determined using equation \ref{eq:triple_based_entity_similarity}.

\begin{equation}
    Pr^{ent.}_{triple}(e_1, e_2) = 1 - \prod_{M_{t(e_1, e_2)}} (1 - Pr^{triple}(t_1, t_2))
    \label{eq:triple_based_entity_similarity}
\end{equation}

\newcounter{obj_type_sim_counter}
\newcommand\rownumber{\stepcounter{obj_type_sim_counter}\arabic{obj_type_sim_counter}}

\begin{table}
    \centering
    \caption{Methods used to calculate the similarity between objects}
    \label{tab:object_type_similarity}
        \begin{tabular}{|c|c|p{0.6\columnwidth}|} \hline
            Number & Object combination & Calculation method \\ \hline
            \ref{tab:object_type_similarity}.\rownumber & entity-entity & Average similarity between entities (Eq. \ref{eq:entity_avg}) \\ \hline
            \ref{tab:object_type_similarity}.\rownumber & entity-string & Extract label from the entity and apply fuzzy string similarity \\ \hline
            \ref{tab:object_type_similarity}.\rownumber & entity-categorical & Extract label from the entity and use category match \\ \hline
            \ref{tab:object_type_similarity}.\rownumber & categorical-categorical & Use string-categorical and choose the highest combination \\ \hline
            \ref{tab:object_type_similarity}.\rownumber & categorical-string & Search for the most similar category value. If it matches, return their similarity. 0.0, otherwise \\ \hline
            \ref{tab:object_type_similarity}.\rownumber & number-number & Scaled euclidean distance \\ \hline
            \ref{tab:object_type_similarity}.\rownumber & number-string & Highest between number-number method on extract numbers from string or edit- distance \\ \hline
            \ref{tab:object_type_similarity}.\rownumber & number-date & Convert date to timestamp and use number-number \\ \hline
            \ref{tab:object_type_similarity}.\rownumber & date-date & Convert both to timestamp and use number-number \\ \hline
            \ref{tab:object_type_similarity}.\rownumber & date-string & Try to extract date from string and use date-date \\ \hline
            \ref{tab:object_type_similarity}.\rownumber & string-string & Calculate fuzzy string similarity \\ \hline
        \end{tabular}
\end{table}

Note that Equation \ref{eq:triple_matching} calculates the similarity between entities by using the subject when the triple represents an attribute and both the subject and object when the triple represents a relationship. Additionally, we use the similarity between triples to inform the entity similarity, as shown in Equation \ref{eq:triple_based_entity_similarity}, creating an interdependence between these equations. To address this, we take the average of the label similarity and triple similarity between entities to compute the overall triple similarity, as outlined in Equation \ref{eq:entity_avg}. When an entity pair lacks either label or triple similarity, we apply a default value of 0.5 for the triple similarity. We use the label similarity calculated in the subsection \ref{subsect:label_matching} for the predicates. The objects can be literal, such as a string and number, or another entity, so we employ different methods for calculating similarity, as summarized in Table \ref{tab:object_type_similarity}.

\begin{equation}
    Pr^{ent.}(e_1 \equiv e_2) = \frac{Pr^{ent.}_{label}(e_1, e_2) + Pr^{ent.}_{triple}(e_1, e_2)}{2}
    \label{eq:entity_avg}
\end{equation}

Consider the example from figure \ref{fig:high-functionality} and the result obtained for triple similarity in the equation \ref{eq:high_triple_matching}. If the entities \textit{Tønsberg} and \textit{New Asgard} only have this pair of triples, we would have entity similarity based on triple as 0.30 (Eq. \ref{eq:high_entity_sim}). The label similarity between the entities is 0.52. We can obtain the entity similarity using the equation \ref{eq:entity_avg} obtaining the value 0.41 (Eq. \ref{eq:high_entity_sim_final}).

\begin{equation}
    Pr^{ent.}_{triple}(e_1, e_2) = 1 - (1 - 0.30) = 0.30
    \label{eq:high_entity_sim}
\end{equation}

\begin{equation}
    Pr^{ent.}(e_1 \equiv e_2) = \frac{0.52 + 0.30}{2} = 0.41
    \label{eq:high_entity_sim_final}
\end{equation}

Consider the example illustrated in Figure~\ref{fig:low-functionality}, where the goal is to compute the similarity between two entities, \textit{A} and \textit{A’}. For simplicity, assume that each entity is associated with exactly two triples. The predicates \textit{dbo:starring} and \textit{mcu:starring} have a predicate-level similarity of 0.9. Their respective functionality and inverse functionality values are as follows: \textit{dbo:starring} has a functionality of 0.26 and an inverse functionality of 0.33, while \textit{mcu:starring} has a functionality of 0.24 and an inverse functionality of 0.30.

Assume further that the entities \textit{Spider-Man: Homecoming} have a similarity of 0.90, and that the pair \textit{The Devil All the Time (film)} and the string \textit{The Devil All the Time} have a fuzzy similarity of 0.80. As the similarity between \textit{A} and \textit{A’} is initially undefined, we set it to a default value of 0.5.

We consider the following triple pairs:

\begin{itemize}
    \item $t_1$ = (A, \textit{dbo:starring}, \textit{Spider-Man: Homecoming}) and $t’_1$ = (A’, \textit{mcu:starring}, \textit{Spider-Man: Homecoming})
    \item $t_2$ = (A, \textit{dbo:starring}, \textit{The Devil All the Time (film)}) and $t’_2$ = (A’, \textit{mcu:wikilink}, \textit{The Devil All the Time})
\end{itemize}

In practice, the cross-pairs $(t_1, t’_2)$ and $(t_2, t’_1)$ would also be considered, but for the sake of simplicity, we assume their similarity is 0.0.

Using Equation~\ref{eq:fun_triples}, the triple similarity based on functionality is computed as in the equation \ref{eq:lower_fun_triple}.
\begin{equation}
    \begin{split}
        Pr^{\text{triple}}_{\text{fun}}(t_1, t'_1) = 0.5 \times 0.9 \times 0.26 \times 0.24 \times 0.90 = 0.03 \\
        Pr^{\text{triple}}_{\text{fun}}(t_2, t'_2) = 0.5 \times 0.9 \times 0.26 \times 0.24 \times 0.80 = 0.03
    \end{split}
    \label{eq:lower_fun_triple}
\end{equation}

Similarly, using Equation~\ref{eq:invfun_triples}, the triple similarity based on inverse functionality is represented in the equation \ref{eq:lower_inv_fun_triple}.

\begin{equation}
    \begin{split}
        Pr^{\text{triple}}_{\text{invFun}}(t_1, t'_1) = 0.5 \times 0.9 \times 0.33 \times 0.30 \times 0.90 = 0.04 \\
        Pr^{\text{triple}}_{\text{invFun}}(t_2, t'_2) = 0.5 \times 0.9 \times 0.33 \times 0.30 \times 0.80 = 0.04
    \end{split}
    \label{eq:lower_inv_fun_triple}
\end{equation}

Using Equation~\ref{eq:triple_matching}, the overall triple similarity is computed as in the equation \ref{eq:lower_triple}.

\begin{equation}
    \begin{split}
        Pr^{\text{triple}}(t_1, t'_1) = 1 - (1 - 0.03)(1 - 0.04) = 0.07 \\
        Pr^{\text{triple}}(t_2, t'_2) = 1 - (1 - 0.03)(1 - 0.04) = 0.07
    \end{split}
    \label{eq:lower_triple}
\end{equation}

Finally, the entity similarity is calculated as in the equation \ref{eq:lower_triple_sim}.

\begin{equation}
    Pr^{\text{ent.}}_{\text{triple}}(A, A') = 1 - (1 - 0.07)(1 - 0.07) = 0.14
    \label{eq:lower_triple_sim}
\end{equation}

This example demonstrates that when only a few triples with moderate similarity are available, the overall entity similarity remains relatively low and may decrease in successive iterations.

Next, consider a different scenario in which the entity pair \textit{(A, A’)} is associated with 10 triples, all having similar objects but low functionality and inverse functionality—such as repeated instances of the triple pair \textit{(A, \textit{dbo:starring}, \textit{Spider-Man: Homecoming})} and \textit{(A’, \textit{mcu:starring}, \textit{Spider-Man: Homecoming})}. Suppose that each of these triple pairs has a similarity of 0.07. The resulting entity similarity is then computed as in the equation \ref{eq:multiple_triples_ent_sim}.

\begin{equation}
    Pr^{\text{ent.}}_{\text{triple}}(A, A') = 1 - (1 - 0.07)^{10} = 0.52
    \label{eq:multiple_triples_ent_sim}
\end{equation}

This case illustrates that even when individual triple similarities are low, a high number of such triples can collectively yield a higher overall entity similarity score.

\subsection{Comparison between PARIS and Full Triple Matcher (FTM)} \label{subsect:comparison}

Our model builds upon the concept of functionality introduced by the PARIS algorithm \cite{suchanek2011}. However, unlike PARIS, which only considers that similar objects imply similar subjects, FTM treats both sides of a triple—subjects and objects—as mutually informative. That is, similar objects suggest similar subjects, and vice versa.

In terms of predicate alignment, PARIS relies on instance-based similarity, whereas our method incorporates both label similarity and semantic similarity computed using BERT. We argue that predicates should be considered similar if they convey semantically equivalent meanings, while PARIS determines similarity based on overlapping object usage. Although PARIS can identify similar object pairs that support entity alignment, this may lead to the generation of semantically unrelated triple pairs. Therefore, we restrict predicate matching to cases supported by label-based semantic similarity.

FTM is also capable of matching semantically similar objects, whereas PARIS relies on exact matches. Although exact matching can reduce computational complexity by limiting the number of candidate pairs, it also restricts the ability to identify near matches that are not lexically identical. Additionally, we account for differences in object types across KGs. For example, a predicate may point to an entity in one KG, while in another, it may reference a literal such as a string representing a name.

From an algorithmic perspective, our model supports multiple input formats, including \textit{TTL}, \textit{RDF/XML}, and SPARQL endpoints. The latter is particularly useful for large-scale knowledge graphs. In contrast, PARIS is limited to the N-Triples format and maintains all potential entity and attribute pairs in matrices or hash tables. While this design facilitates faster execution and parallel processing for smaller datasets, it also demands significant memory resources when applied to large KGs.

Moreover, our method generates triple-level alignments, which can enhance the interpretability of the entity matching process and provide insights into the entity alignment results.
\section{Algorithm} \label{section:implementation}

In this subsection, we introduce the implemented algorithm, which aims to map triples between two distinct knowledge graphs (KGs), denoted as $\mathcal{G}_1$ and $\mathcal{G}_2$. The algorithm outputs a set of mapped triples, $\mathcal{M}_t$.

\begin{figure}
    \centering
    \includegraphics[width=\columnwidth]{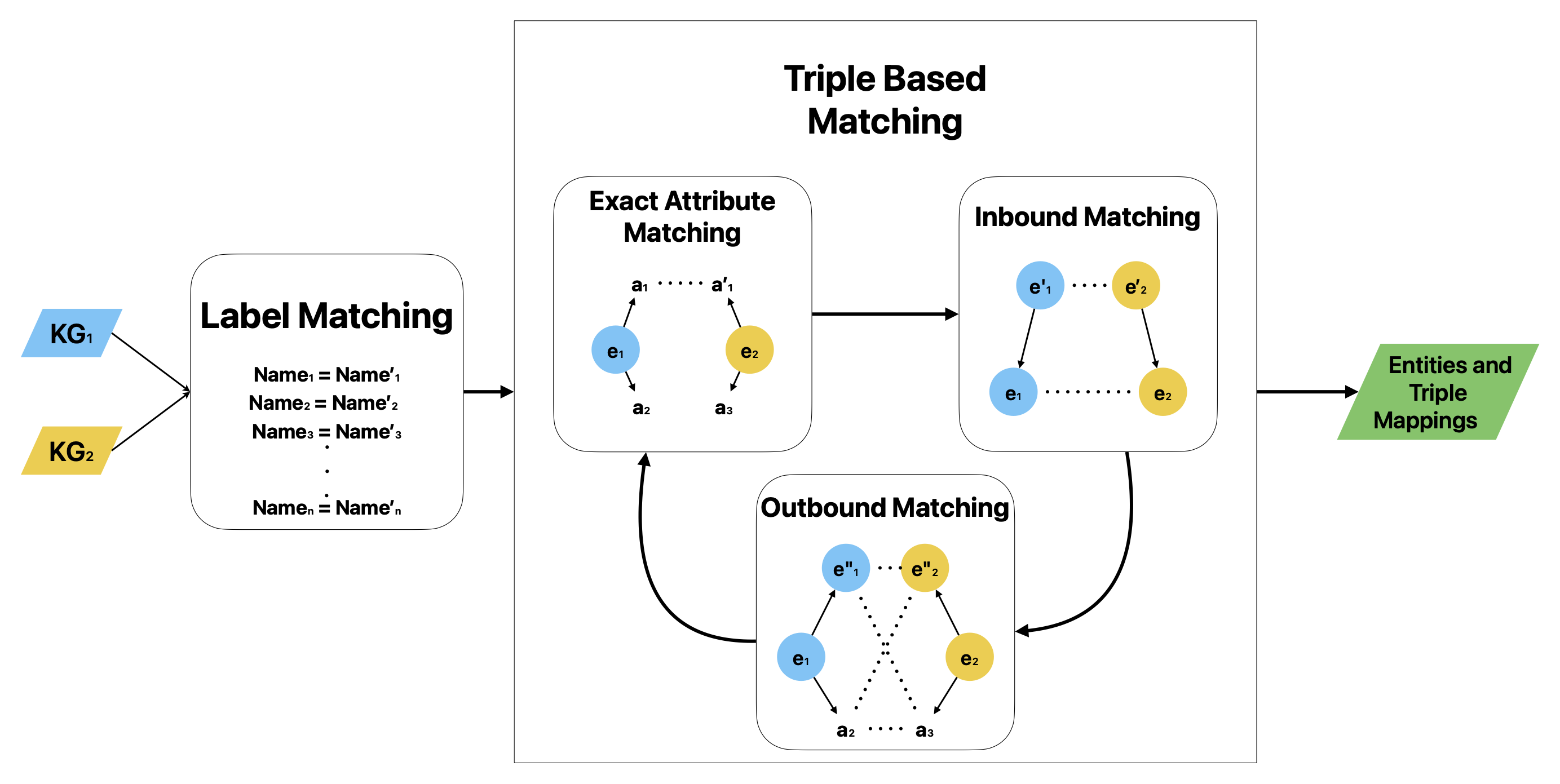}
    \Description[Full Triple Matcher Flowchart]{Full Triple Matcher Flowchart}
    \caption{Full Triple Matcher Flowchart}
    \label{fig:flowchart}
\end{figure}

Figure \ref{fig:flowchart} represents the structure used to implement the proposed algorithm. The process begins by calculating the similarity between entity and predicate labels in the two KGs (Subsect. \ref{subsect:label_matching}). Next, entity similarity is computed based on the triples associated with each entity (Subsect \ref{subsect:triple_based_matching}). Exact attribute matching identifies triples with identical literal values (Subsubsect. \ref{subsect:exact_attribute_matching}). In contrast, inbound and outbound matching retrieves triples from entity pairs to search for similar triples and expand the pairs in subsubsection \ref{subsect:inbound_matching} and \ref{subsect:outbound_matching}, respectively.

These three steps—exact attribute matching, inbound matching, and outbound matching—are iterated to refine the matching list, ultimately producing the final set of aligned entities and triples. Figure \ref{fig:flowchart} represents the process used in this paper. 

\subsection{Label Matching} \label{subsect:label_matching}

This step uses the URI and the elements' labels to calculate their confidence. The algorithm receives two KGs $\mathcal{G}_1$ and $\mathcal{G}_2$ and returns a set of mappings $m_p = (p_1, p_2, c) \in \mathcal{M_P}$ and $m_{label} = (e_1, e_2, c_l) \in \mathcal{M}_{label}$, where $c \in \mathbb{R}$. The confidence value is the highest of the following methods, depending on the similarity between the two elements. First, if the URI matches, the confidence is 1.0. 
Second, if the labels have an exact match, the confidence is 0.9. Third, the labels are preprocessed by removing parenthesis and punctuation, splitting camel case and numbers, replacing underscore with spaces, and removing unnecessary spaces. Then, if the normalized label matches, the confidence is 0.8. Fourth, the algorithm eliminates stopwords using the NLTK library \cite{steven2004} and returns 0.7 if they match.

The last two methods are used if there are less than 1,000 elements. A cross-product is used between all pairs of labels to calculate the confidence. Fuzzy similarity uses the weighted ratio from the RapidFuzz library \cite{bachmann2023} and multiplies the result by 0.7. The BERT similarity uses the BERT model \cite{devlin2018bert} to convert the label into a vector, then calculates the similarity between two vectors using cosine similarity and multiplies by 0.7. We used a pre-trained English BERT model (bert-base-uncased) with 110 million parameters obtained from Hugging Face.\footnote{\url{https://huggingface.co/google-bert/bert-base-uncased}}.
We chose this model because, in our testing, larger models had only a minor impact on the final performance, as it is used solely as an initial step in the entity and triple matching process. Therefore, we selected a small yet high-performance model, such as BERT.
Table \ref{tab:label_sim} summarized the label similarity steps and returned values.

\begin{table}[t]
    \centering
    \caption{Label similarity values}
    \begin{tabular}{|c||c|} \hline
        Similarity method & Similarity value \\ \hline
        URI exact match & 1.0 \\ \hline
        Label exact match & 0.9 \\ \hline
        Normalized label match & 0.8 \\ \hline
        Remove stopwords & 0.7 \\ \hline
        Fuzzy match & $0.7 \times fuzzySimilarity$ \\ \hline
        BERT Similarity & $0.7 \times BERTSimilarity$ \\ \hline
    \end{tabular}
    
    \label{tab:label_sim}
\end{table}

\subsection{Triple Based Matching} \label{subsect:triple_based_matching}

This step focuses on matching triples that provide a more comprehensive description of an entity, thus enhancing the accuracy of entity matching. Subsection \ref{subsubsect:algorithm} outlines the algorithm implemented in FTM, while subsections \ref{subsect:exact_attribute_matching}, \ref{subsect:inbound_matching}, and \ref{subsect:outbound_matching} examine the substeps of the algorithm in greater detail.

\subsubsection{Algorithm} \label{subsubsect:algorithm}

The algorithm \ref{alg:our_paris} describes a simplified version of our implementation. The algorithm takes two knowledge graphs, denoted as $\mathcal{G}_1$ and $\mathcal{G}2$, along with the mapping $\mathcal{M}_{label}$ between entities $e_1 \in \mathcal{E}_1$ and $e_2 \in \mathcal{E}_2$, computed during the label similarity step (subsection \ref{subsect:label_matching}) as input. Initiated by loading literals from each KG and establishing the set of common literals between them (line 1), the algorithm loads triples where the object value aligns with the common literals (lines 2-3).

Lines 4 to 6 initialize the counter and the variables $M_{t}$ and $M^{ent}_{t}$, which store mappings between triples and entities produced by our algorithm. Line 8 calculates the average of the label confidence and triple similarity. Since only label confidence is available in the initial iteration, its value is weighted by multiplying it by 0.5. In later iterations, if only triple similarity is present for an entity pair, we retrieve the labels of each entity and use BERT similarity as the label confidence. Lines 9 and 10 identify triples with common literals to create mappings between triples with identical objects (Subsect. \ref{subsect:exact_attribute_matching}). Lines 11 to 22 load all triples from previously matched entity pairs to establish mappings between triples. Specifically, lines 11 to 16 retrieve inbound triples (Subsect. \ref{subsect:inbound_matching}), while lines 17 to 22 retrieve outbound triples (Subsect. \ref{subsect:outbound_matching}). Lines 23 to 25 extract all entity pairs generated in the triple mappings and compute their similarity using Equation \ref{eq:triple_based_entity_similarity}. The triple-matching process iterates further if the matched source entities increase by over 10\% or if the top entity pairs shift by more than 10\% (lines 27-28). To ensure efficiency, we limited the iterations to ten; however, the process consistently converged after four iterations in all tested cases.

The divergence between triples is not used in entity similarity calculations. Therefore, we only compute triple divergence after the matching process. The algorithm selects the highest matching pair for each entity, retrieves the associated triples, and performs a cross-product. Finally, it calculates the divergence for each pair of triples using equation \ref{eq:triple_div_matching}.

\begin{algorithm}
\caption{Full Triple Matcher implementation}\label{alg:our_paris}
\KwData{$\mathcal{G}_1 =(\mathcal{E}_1, \mathcal{P}_1, \mathcal{L}_1, \mathcal{T}_1), \mathcal{G}_2 =(\mathcal{E}_2, \mathcal{P}_2, \mathcal{L}_2, \mathcal{T}_2), (e_1, e_2, c) \in \mathcal{M}_{label}$}
\KwResult{$(e_1, e_2, conf) \in \mathcal{M}_e$}
$L \gets \mathcal{L}_1 \cap \mathcal{L}_2$\;
$T_{literal1} \gets \{(e_1, p_1, o) \mid t = (e_1, p_1, o) \in \mathcal{T}_1, o \in L\}$\;
$T_{literal2} \gets \{(e_2, p_2, o) \mid t = (e_2, p_2, o) \in \mathcal{T}_2, o \in L\}$\;
$counter \gets 0$\;
$M_{t} \gets \emptyset$\;
$M^{ent}_{t} \gets \emptyset$\;
\While{$counter < 10$}{
    $M_e \gets \{(e_1, e_2, c) \mid ((e_1, e_2, c_{l}) \in M_{label} \lor (e_1, e_2, c_{t}) \in M^{ent}_{t}), c = 0.5 * c_{l} + 0.5 * c_{t}\}$\;
    \ForEach{$(t_1, t_2) \mid t_1 = (e_1, p_1, o_1) \in T_{literal1}, t_2 = (e_2, p_2, o_2) \in T_{literal2}, o_1 = o_2$}{
        $M_{t} \cup \{(t_1, t_2, conf_{t}) \mid conf = Pr^{triple}_{exactMatch}(t_1, t_2)\}$\;
    }
    \ForEach{$(e_1, e_2, conf) \in M_e$}{
        $T_1 \gets \{t_1 = (s_1, p_1, o_1) \in \mathcal{T}_1 \mid o_1 = e_1\}$\;
        $T'_2 \gets \{t_2 = (s_2, p_2, o_2) \in \mathcal{T}_1\mid o_2 = e_2\}$\;
        $M_{inbound} \gets T_1 \times T_2$\;
        \ForEach{$(t_1, t_2) \in M_{inbound}$}{
            $M_{t} \cup {(t_1, t_2, conf_{t})} \mid conf = Pr^{triple}_{inbound}(t_1, t_2, conf_{t})$\;
        }
    }
    \ForEach{$(e_1, e_2, conf) \in M_e$}{
        $T_1 \gets \{t_1 = (s_1, p_1, o_1) \in \mathcal{T}_1 \mid s_1 = e_1\}$\;
        $T_2 \gets \{t_2 = (s_2, p_2, o_2) \in \mathcal{T}_1\mid s_2 = e_2\}$\;
        $M_{outbound} \gets T_1 \times T_2$\;
        \ForEach{$(t_1, t_2) \in M_{outbound}$}{
            $M_{t} \cup {(t_1, t_2, conf_{t})} \mid conf_{t} = Pr^{triple}_{outbound}(t_1, t_2)$\;
        }
    }

    \ForEach{$(e_1, e_2) \mid (t_1, t_2, conf) \in M_t \wedge ((e_1 = s_1 \wedge e_2 = s_2) \vee (e_1 = o_1 \wedge e_2 = o_2))$}{
        $M_{t(e_1, e_2)} \gets \{ (t_1, t_2, conf) \in M_t \wedge ((e_1 = s_1 \wedge e_2 = s_2) \vee (e_1 = o_1 \wedge e_2 = o_2))\}$\;
        $M^{ent}_{t} \cup \{(e_1, e_2, conf) \mid conf = Pr^{ent}_{triple}(M_{t(e_1, e_2)})\}$\;
    }
    $counter \gets counter + 1$\;
    \uIf{$EarlyStopCondition$}{$break$\;}{}
}
\end{algorithm}

\subsubsection{Exact Attribute Matching} \label{subsect:exact_attribute_matching}

This step aims to expand the set of entity pairs by identifying exactly matching attributes. We extract all literal values from the objects, identify equal ones, and create mappings between triples that share the same object to achieve this. Formally, given two triples, $t_1 = (e_1, p_1, a)$ and $t_2 = (e_2, p_2, a’)$, where $a = a’$, we compute the probability of $t_1$ and $t_2$ being similar, as described in equation \ref{eq:exact_attribute_matching}. In this case, no similarity calculation is required between the objects since they are identical.

\begin{multline}
    Pr^{triple}_{exactMatch}(t_1, \equiv t_2) = 1 - (1 - Pr^{ent.}(e_1 \equiv e_2) \times Pr^{pred.}(p_1 \equiv p_2) \times fun(p_1) \times fun(p_2)) \\ \times (1 - Pr^{ent.}(e_1 \equiv e_2) \times Pr^{pred.}(p_1 \equiv p_2) \times invFun(p_1) \times invFun(p_2))
    \label{eq:exact_attribute_matching}
\end{multline}

\subsubsection{Inbound Matching} \label{subsect:inbound_matching}

In this step, we utilize the set of entity pairs for which similarity has already been calculated to match the inbound triples—those where the entity appears as the object. For each entity $e_1$, we rank its similarity to target entities and select the top 10 pairs $(e_1, e_2)$ with the highest similarity. We then retrieve all triples where these entities are objects, generate a set of triple mappings through a cross-product, and compute their similarity.

Formally, for two entities $e_1$ and $e_2$, with their average probability $Pr^{ent.}(e_1 \equiv e_2)$ already determined, we extract the triples $t’_1 = (e’_1, p_1, e_1) \in T’_1$ and $t’_2 = (e’_2, p_2, e_2) \in T’_2$. We then create the set of triple mappings $M{t(e_1, e_2)} = T’_1 \times T’_2$ and compute the similarity between these triples using equation \ref{eq:relationship_matching}.

\begin{multline}
    Pr^{triple}_{inbound}(t_1 \equiv t_2) = 1 - (1 - Pr^{ent.}(e_1 \equiv e_2) \times Pr^{pred.}(p_1 \equiv p_2) \times fun(p_1) \times fun(p_2) \times Pr^{ent.}(e'_1 \equiv e'_2)) \\ \times (1 - Pr^{ent.}(e_1 \equiv e_2) \times Pr^{pred.}(p_1 \equiv p_2) \times invFun(p_1) \times invFun(p_2) \times Pr^{ent.}(e'_1, e'_2))
    \label{eq:relationship_matching}
\end{multline}

\subsubsection{Outbound Matching} \label{subsect:outbound_matching}

In this step, we use the set of entity pairs with pre-calculated similarity scores to match the outbound triples - those where the entity appears as the subject. We focused solely on exact object matches in subsection \ref{subsect:exact_attribute_matching}. However, objects may convey similar information in different formats, such as addresses or movie lengths. Additionally, we match entities and literals, as certain relationships may be represented as an entity in one KG but as a string in another.

For each entity $e_1$, we rank its similarity to the target entities and select the top 10 pairs $(e_1, e_2)$ with the highest similarity. We then retrieve all triples where the entities are subject, map them as the cross-product of retrieved triples, and compute their similarity. Formally, given two entities are given $(e_1, e_2)$, we retrieve the triples $t'_1 = (e_1, p_1, o_1) \in T'_1$  and $t'_2 = (e_2, p_2, o_2) \in T'_2$. The set of triple mappings $M_{t(e_1, e_2)}$ is then created as $T'_1 \times T'_2$, and the similarity between triples is computed as described in equation \ref{eq:object_matching}. The similarity between objects, $Pr(o_1, o_2)$, is determined based on the various object types listed in Table \ref{tab:object_type_similarity}.

\begin{multline}
    Pr^{triple}_{outbound}(t_1 \equiv t_2) = 1 - (1 - Pr^{entity}(e_1 \equiv e_2) \times Pr^{pred.}(p_1 \equiv p_2) \times fun(p_1) \times fun(p_2) \times Pr^{obj.}(o_1 \equiv o_2)) \\ \times (1 - Pr^{ent.}(e_1 \equiv e_2) \times Pr^{pred.}(p_1 \equiv p_2) \times invFun(p_1) \times invFun(p_2) \times Pr^{obj.}(o_1 \equiv o_2))
    \label{eq:object_matching}
\end{multline}
\section{Experiment} \label{section:experiment}

In this section, we outline the experiments conducted to evaluate FTM. We begin by introducing the dataset, metrics, and computer specifications utilized in the experiment (Subsection \ref{subsect:experimental_setup}), followed by a presentation of the results obtained (Subsection \ref{subsect:result}).

\subsection{Experimental setup} \label{subsect:experimental_setup}

This subsection outlines the dataset employed to evaluate the entity matching phase, the entity matching techniques utilized for comparison against FTM, and the relevant metrics (Subsect. \ref{subsect:dataset_entity_matching}). Following this, we elucidate the methodology for acquiring the dataset utilized in formulating the dataset for the triple matching phase and the metric employed (Subsect. \ref{subsect:dataset_triple_matching}).

\subsubsection{Entity matching} \label{subsect:dataset_entity_matching}

We evaluated our entity-matching method using the KG track of the OAEI 2023, based on datasets from the DBkWik project \cite{hertling2020dbkwik}. DBkWik extends the DBpedia Extraction Framework to construct KGs from general Wiki community pages using MediaWiki XML dumps \cite{hertling2018}. Each Wiki page becomes an entity, and each infobox entry forms a triple: links generate entity relations, while plain text yields literals. Manual mappings between Wiki elements and DBpedia ontology terms guide the extraction, but are impractical for generic dumps. Consequently, DBkWik treats each infobox key as a property and each infobox type as a class. Table \ref{tab:unsupervisedKgStatistics} summarizes the dataset’s KGs. The gold standard was manually constructed by experts at the schema level, and entity links were derived from hyperlinks found in the external links section of original Wiki pages \cite{hertling2020}.

\begin{table}[htp]
    \centering
    \footnotesize
    \caption{Gold standard statistics for unsupervised methods \cite{hertling2020}}
        \begin{tabular}{|c|c|c|c|} \hline
        \multirow{2}{4em}{Mapping}  & Entity Matches & \multicolumn{2}{|c|}{Triple Matches} \\ \cline{2-4}
        & {Total} & {Compatible} & {Divergent} \\ \hline
        {SWW-SWG} & 1,096 & 117 & 15\\ \hline
        {SWW-TOR} & 1,358 & 1,131 & 889 \\ \hline
        {MCU-MDB} & 1,654 & 0 & 0\\ \hline
        {MAL-MBT} & 9,296 & 4,765 & 3,047\\ \hline
        {MAL-STX} & 1,725 & 752 & 540\\ \hline
        {MAL-DBpedia} & 2,178 & 162 & 81 \\ \hline
        {SWW-DBpedia} & 2,193 & 255 & 73 \\ \hline
        \end{tabular}
    \label{tab:gsStatistics}
\end{table}

Table \ref{tab:gsStatistics} elucidates the dataset employed in our experimentation. Our comparative analysis involved the utilization of BaselineAltLabel, LogMap, and PARIS alongside our proposed approach. BaselineAltLabel, provided by the OAEI Knowledge Graph Track organization, employs labels (\textit{rdfs:label}) and alternative labels (\textit{skos:altLabel}) of entities to seek an exact match between them. LogMap has the second-best performance among competitors in the OAEI Knowledge Graph Track 2023, only losing to the BaselineAltLabel \cite{pour2023a}. The PARIS algorithm was assessed using the original code accessible in \cite{suchanek2011} and explained in Section \ref{subsection:paris}. 

We also compared FTM with the baseline for the Memory Alpha KG, Star Wars Wiki KG, and DBpedia (2022-09) to test the performance using a real-size KG and matching two KGs with different sizes. The table shows the size of the KGs used in this test. The matching between both datasets was extracted from the Gollum dataset \cite{hertling2022gollum}. Expanding on the work of DBkWik, Gollum presents an approach designed explicitly for large-scale matching of KGs from diverse sources. While it utilizes the same KG extraction method as DBkWik, Gollum enhances entity matching by incorporating transitive closure, thereby significantly increasing the number of identified correspondences. This approach has created a gold standard dataset with 4,149 KGs and a scalability testing dataset with 307,466 KGs. We used the KGs extracted from the DBkWik and Gollum datasets as a community-based KG to test FTM.

The DWY-NB dataset was chosen from \cite{zhang2022benchmark} to compare FTM with supervised methods. 
It aims to tackle two primary challenges: bijection and name variety. Bijection refers to a dataset containing KGs where nearly all entities have a one-to-one alignment. However, real-world scenarios often involve KGs extracted from diverse sources or domains, leading to entities without alignment. Therefore, it is expected to include such entities. Name variety is also crucial in developing a dataset where entity alignment cannot be achieved solely by matching their labels. To address these challenges, the authors removed 25\% of the alignments in each KG and chose a different name for entities with multiple name attributes. Through these procedures, they generated a dataset where 50\% of entities lack alignments and 36\% of aligned entities have distinct names.

We selected AttrE \cite{trisedya2019entity}, NMN \cite{wu-2020-neighborhood} and AutoAlign \cite{zhang2023autoalign} for comparison with the unsupervised methods and FTM. These methods were chosen because they were identified as the top translational and GCN-based approaches for the DWY-NB dataset \cite{zhang2022benchmark}. AttrE embeds attribute values and employs a translational model to understand the KG structure embedding by translating entities into their attributes \cite{trisedya2019entity}. On the other hand, NMN utilizes GCN to capture the KG structure and incorporates an attention mechanism to comprehend the neighborhood difference, aiming to address the structural heterogeneity between KGs \cite{wu-2020-neighborhood}. AutoAlign is an unsupervised KG matcher that creates a predicate-proximity graph using LLM \cite{zhang2023autoalign}. The results for the supervised methods were extracted from \cite{zhang2022benchmark}. They executed the experiment in an Intel(R) Xeon(R) CPU E5-2650 v4, 128GB main memory, and an Nvidia Tesla GPU with 32GB memory. We executed the unsupervised methods on a MacBook Pro with Apple M1 Max and 64GB of main memory. For entity matching with DBpedia, we tested the matchers using a server with 40 CPUs Intel(R) Xeon(R) Silver 4210R, and 754 Gigabytes of RAM.

We evaluated the performance of the entity alignment methods for both supervised and unsupervised approaches using the Hit@1 and Hit@10 metrics. These metrics were selected to measure the effectiveness of each technique in correctly identifying the corresponding entity for alignment. A higher Hit@k score indicates better alignment performance.

We also evaluated precision, recall, and F-measure for the unsupervised method. The evaluation was conducted under two key assumptions in the dataset: one-to-one matching and the open-world assumption. The one-to-one matching assumption stipulates that each entity corresponds to at most one entity in the other knowledge graph (KG). Therefore, we select the target entity pair with the highest similarity score for each source entity. In cases where multiple target entities share the highest similarity score, all such pairs are selected.

The open-world assumption acknowledges that the gold standard does not necessarily contain all correct entity correspondences. Accordingly, we only evaluate entity pairs that are present in the gold standard. Let $P$ denote the set of positive matches (i.e., all entity pairs included in the gold standard). True positive matches ($TP$) are entity pairs where the source and target entities are correctly matched according to the gold standard. False positive matches ($FP$) refer to entity pairs in which at least one entity is present in the gold standard but is incorrectly linked to a different entity. False negative matches ($FN$) are entity pairs that exist in the gold standard but are not predicted by the evaluated model. Precision, recall, and F-measure are calculated according to Equations \ref{eq:precision}, \ref{eq:recall}, and \ref{eq:f-measure}. A higher number indicates better performance for all metrics.

\begin{equation}
    Precision = \frac{TP}{TP + FP}
    \label{eq:precision}
\end{equation}

\begin{equation}
    Recall = \frac{TP}{P}
    \label{eq:recall}
\end{equation}

\begin{equation}
    F-measure = \frac{2 \times TP}{2 \times TP + FP + FN}
    \label{eq:f-measure}
\end{equation}

\subsubsection{Triple matching} \label{subsect:dataset_triple_matching}

To evaluate our triple-matching approach, we adapted the KG track within the OAEI competition 2023 from entity matching to create the set of mappings between triples and manually evaluated each pair of triples. We also used the gold standard for predicate and entity pairs from the OAEI competition to create the dataset. The gold standard has an open-world assumption, so we only consider the pairs that are included in the gold standard. First, we selected the predicates in the gold standard with functionality above 0.8. Second, we extracted all triples with entities in the gold standard and predicates in the filtered set. Finally, we created the list of mappings between triples with corresponding entity and predicate pairs in the gold standard. In the manual evaluation, if the objects are both entities, we define them as compatible if they are in the gold standard, divergent if a mapping exists to another entity, and excluded when both are not in the gold standard. If one of the objects is literal, we check if they have the same data type or have the same label as another one that is an entity. Minor differences in typos, such as lower or upper case, were considered compatible. Some synonyms such as \textit{pale}, \textit{white}, and \textit{fair} were also considered compatibles. For numbers, we defined them as compatible if the difference is less than 10\% or with a difference smaller than 10 when both values are less than 100. For dates, if both objects refer to the same date, we consider them as compatible even if the format is different.

We evaluated the triple-matching approach using precision, recall, and F-measure, as outlined in Subsection~\ref{subsect:dataset_entity_matching}. In this context, a true positive is defined as a triple pair that exists in the gold standard and is correctly identified by FTM. A false positive refers to a triple pair that exists in the gold standard but is incorrectly classified, while a false negative denotes a gold standard triple pair that was not predicted by FTM.

The codes, dataset, and results created for this experiment are available on GitHub. \footnote{\url{https://github.com/eitiyamamoto/Full-Triple-Matcher}}

{\small
\begin{table}[htp]
    \centering
    \caption{Statistics of knowledge graphs for unsupervised methods \cite{hertling2020}}
    \footnotesize
    \begin{tabular}{|p{0.4\columnwidth}||c|c|c|}\hline
       Source & \# Entities & \# Properties & \# Classes \\ \hline
        {Star Wars Wiki (SWW)}  & {145,033} & {700} & {269} \\ \hline
       {The Old Republic Wiki (TOR)} & {4,180} & {368} & {101} \\ \hline
        {Star Wars Galaxies Wiki (SWG)} & {9,634} & {148} & {67} \\ \hline
        {Marvel Database (MDB)}  & {210,996} & {139} & {186} \\ \hline
       {Marvel Cinematic Universe Wiki (MCU)} & {17,187} & {147} & {55} \\ \hline
        {Memory Alpha (MAL)} & {45,828} & {325} & {181} \\ \hline
        {Star Trek Expanded Universe (STX)}  & {13,426} & {202} & {283} \\ \hline
        {Memory Beta (MBT)} & {51,323} & {423} & {240} \\ \hline
        {DBpedia (2022-09)} & {18,996,572} & {54,655} & {483,445} \\ \hline
    \end{tabular}
    \label{tab:unsupervisedKgStatistics}
\end{table}}

\begin{table}[htp]
    \centering
    \caption{Statistics of knowledge graphs for supervised methods}
    \footnotesize
    \begin{tabular}{|c|c|c||c|c|c|}\hline
       Dataset & Matches & Source & \# Entities & \# Properties & \# Classes \\ \hline
       \multirow{2}{*}{DW-NB} & \multirow{2}{*}{50,000} & {DBpedia}  & {84,911} & {545} & {0} \\ \cline{3-6}
       & & {Wikidata} & {86,116} & {700} & {0} \\ \hline
       \multirow{2}{*}{DY-NB} & \multirow{2}{*}{15,000} & {DBpedia}  & {58,858} & {211} & {0} \\ \cline{3-6}
       & & {Yago} & {60,228} & {91} & {0} \\ \hline
       \multirow{2}{*}{DW-300k} & \multirow{2}{*}{150,000} & DBpedia & 150,000 & 4,380 & 804 \\ \cline{3-6}
       & & Wikidata & 150,000 & 1,551 & 7 \\ \hline
       \multirow{2}{*}{DW-600k} & \multirow{2}{*}{300,000} & DBpedia & 300,000 & 7,208 & 3,337 \\ \cline{3-6}
       & & Wikidata & 300,000 & 2,186 & 12 \\ \hline
    \end{tabular}
    \label{tab:supervisedKgStatistics}
\end{table}

\normalsize



\subsection{Results} \label{subsect:result}


\subsubsection{Entity matching} \label{subsect:results_entity_matching}

The table \ref{tab:result_entity_matching} displays the outcomes derived from unsupervised entity matching methods applied to the dataset referenced in table \ref{tab:gsStatistics}. For the SWW-SWG dataset, both LogMap and FTM achieved identical results for hit@10, while LogMap demonstrated superior performance for hit@1. Regarding the SWW-TOR dataset, FTM exhibited the best hit@1 and hit@10 performance. In the case of MCU-MDB, LogMap, and PARIS showcased the best results for both hit@1 and hit@10. Turning to MAL-MBT, PARIS showed the best hit@1 performance, while FTM excelled in hit@10. Similarly, for MAL-STX, FTM performed better for both hit@1 and hit@10.

LogMap, PARIS, and FTM achieved the best hit@1 results for the two datasets in summarizing the comparisons. Regarding hit@10, FTM demonstrated the best performance across four datasets, while LogMap performed better in two datasets and PARIS in one. For hit@1, three matchers demonstrated similar performances, but FTM outperformed when comparing using hit@10.


FTM consistently achieved the highest or near-highest Hit@1 and Hit@10 scores across most datasets, outperforming the baseline methods in terms of candidate retrieval performance. When we use the highest f-measure to compare, BaselineAltLabel and LogMap have the highest performance. However, FTM improved the precision compared to PARIS and have a competitive result to the baseline approaches.

{
\color{red}
\footnotesize
\begin{table}[t]
    \centering
    \footnotesize
    \caption{Result for unsupervised entity matching dataset}
    \begin{tabular}{|c|c|c|c||c|c|c|c|} \hline
        dataset & matcher & Hit@1 & Hit@10 & Precision & Recall & F-measure & Threshold \\ \hline
        \multirow{4}{5em}{SWW-SWG} & BaselineAltLabel & 0.63 & 0.63 & 0.40 & 0.62 & 0.48 & 0.0 \\ \cline{2-2}
        & LogMap & \textbf{0.83} & \textbf{0.83} & \textbf{0.49} & \textbf{0.69} & \textbf{0.57} & 0.95 \\ \cline{2-2}
        & PARIS & 0.73 & 0.73 & 0.18 & 0.67 & 0.28 & 0.99\\ \cline{2-2}
        & FTM & {0.81} & \textbf{0.83} & 0.36 & 0.62 & 0.45 & 0.90 \\ \hline \hline
        
        \multirow{4}{5em}{SWW-TOR} & BaselineAltLabel & 0.91 & 0.91 & 0.46 & 0.90 & \textbf{0.60} & 0.0 \\ \cline{2-2}
        & LogMap & 0.90 & 0.90 & \textbf{0.49} & 0.78 & \textbf{0.60} & 0.70 \\ \cline{2-2}
        & PARIS & {0.94} & 0.94 & 0.20 & \textbf{0.93} & 0.33 & 0.99 \\ \cline{2-2}
        & FTM & \textbf{0.95} & \textbf{0.96} & 0.40 & 0.90 & 0.55 & 0.90 \\ \hline \hline
        
        \multirow{4}{5em}{MCU-MDB} & BaselineAltLabel & {0.68} & {0.68} & \textbf{0.44} & \textbf{0.68} & \textbf{0.53} & 0.0 \\ \cline{2-2}
        & LogMap & \textbf{0.81} & \textbf{0.81} & 0.34 & 0.46 & 0.39 & 0.99 \\ \cline{2-2}
        & PARIS & \textbf{0.81} & \textbf{0.81} & 0.35 & 0.78 & 0.49 & 0.99\\ \cline{2-2}
        & FTM & 0.71 & 0.80 & 0.39 & \textbf{0.68} & 0.49 & 0.90 \\ \hline \hline
        
        \multirow{4}{5em}{MAL-MBT} & BaselineAltLabel & 0.89 & 0.89 & \textbf{0.62} & 0.89 & \textbf{0.73} & 0.0 \\ \cline{2-2}
        & LogMap & 0.79 & 0.79 & 0.61 & 0.76 & 0.68 & 0.97 \\ \cline{2-2}
        & PARIS & \textbf{0.93} & 0.93 & 0.39 & \textbf{0.92} & 0.55 & 0.99 \\ \cline{2-2}
        & FTM & {0.92} & \textbf{0.97} & 0.60 & 0.89 & 0.72 & 0.94 \\ \hline \hline
        
        \multirow{4}{5em}{MAL-STX} & BaselineAltLabel & 0.92 & 0.92 & 0.49 & \textbf{0.93} & 0.64 & 0.0 \\ \cline{2-2}
        & LogMap & 0.79 & 0.79 & \textbf{0.55} & 0.77 & \textbf{0.65} & 0.97 \\ \cline{2-2}
        & PARIS & {0.93} & 0.93 & 0.25 & \textbf{0.93} & 0.39 & 0.99\\ \cline{2-2}
        & FTM & \textbf{0.94} & \textbf{0.97} & 0.45 & 0.92 & 0.60 & 0.90 \\ \hline

    \end{tabular}
    \label{tab:result_entity_matching}
\end{table}
}

The table \ref{tab:result_entity_matching_real} presents the results of employing unsupervised entity matching techniques on two different datasets: a large-scale general dataset (DBpedia) and a specific dataset (Memory Alpha and Star Wars Wiki). LogMap encountered difficulties loading the entire DBpedia, leading to the inability to generate a mapping set. As a result, we deemed LogMap as not applicable (n/a) for this matching process.

Regarding the Memory Alpha and DBpedia dataset (MAL-DBpedia), FTM performed the best in Hit@1 and Hit@10. As for the Star Wars Wiki and DBpedia dataset (SWW-DBpedia), we executed the PARIS method for over three weeks. However, due to the projected completion time exceeding three months, we terminated the process and classified it as not applicable.
In this dataset comparison, our method delivered superior results for both Hit@1 and Hit@10. In summary, only the BaselineAltLabel method and FTM completed the matching process within a reasonable timeframe. FTM exhibited the best overall performance across the datasets regarding Hit@1 and Hit@10.

Only the BaselineAltLabel and our proposed approach could be executed for this dataset, as LogMap was not possible to run. BaselineAltLabel achieves higher precision and F-measure on the Memory Alpha and DBpedia datasets. In contrast, FTM demonstrates superior precision and F-measure on the Star Wars Wiki and DBpedia datasets. While BaselineAltLabel obtains higher recall on the Star Wars Wiki and DBpedia datasets, our method achieves better recall on the Memory Alpha and DBpedia datasets. These results illustrate the challenge of maintaining a balanced trade-off between precision and recall in this context. Overall, all models exhibit low precision, which can be attributed to the presence of multiple entities with identical names referring to different concepts. Despite this, FTM achieves good performance and maintains a stable hit rate across datasets.

\begin{table}[t]
    \centering
    \footnotesize
    \caption{Result for unsupervised entity matching methods for real-size KGs }
    \begin{tabular}{|c|c|c|c||c|c|c|c||c|} \hline
        dataset & matcher & Hit@1 & Hit@10 & Precision & Recall & F-measure & Threshold & Runtime \\ \hline
        
        \multirow{4}{5em}{MAL-DBpedia} & BaselineAltLabel & {0.69} & 0.69 & \textbf{0.07} & 0.69 & \textbf{0.12} & 0.0 & 2.15 days \\ \cline{2-4}
        & LogMap & n/a & n/a & n/a & n/a & n/a & n/a & n/a \\ \cline{2-4}
        & PARIS & 0.54 & 0.54 & 0.04 & 0.50 & 0.07 & 0.99 & 18.40 hours \\ \cline{2-4}
        & FTM & \textbf{0.75} & \textbf{0.89} & 0.06 & \textbf{0.73} & 0.10 & 0.94 & 10.04 days \\ \hline \hline

        \multirow{4}{5em}{SWW-DBpedia} & BaselineAltLabel & {0.41} & 0.78 & 0.04 & \textbf{0.78} & 0.08 & 0.0 & 4.67 days \\ \cline{2-4}
        & LogMap & n/a & n/a & n/a & n/a & n/a & n/a & n/a \\ \cline{2-4}
        & PARIS & n/a & n/a & n/a & n/a & n/a & n/a & n/a \\ \cline{2-4}
        & FTM & \textbf{0.72} & \textbf{0.83} & \textbf{0.06} & 0.71 & \textbf{0.12} & 0.94 & 6.07 days \\ \hline
    \end{tabular}
    
    \label{tab:result_entity_matching_real}
\end{table}

The table \ref{tab:result_entity_matching_supervised} illustrates the outcomes of utilizing unsupervised and supervised techniques for entity matching on the dataset depicted in the table \ref{tab:supervisedKgStatistics}. The data marked with \textsuperscript{a} were extracted from \cite{zhang2022benchmark}. The data marked with \textsuperscript{b} were extracted from \cite{zhang2023autoalign}. LogMap successfully loaded the dataset, but the datasets do not contain sufficient classes, directly impacting the LogMap method. As a result, the matcher could not generate any mappings, leading us to classify it as not applicable. We successfully reproduced AutoAlign using DW-NB and DY-NB datasets, but the matcher could not complete the task for DW-300k and DW-600k. For this reason, we classified it as not applicable to both datasets.

Supervised entity matching necessitates a training set, with the size of this set influencing performance. Hence, we present the entity matcher results using 10\% and 50\% of entity alignment as the training set (seeds). Our method exhibited the most favorable results for both metrics for all datasets except for DW-NB where PARIS has a better Hit@1 and FTM and AutoAlign tied at Hit@10.

{
\begin{table}[t]
    \centering
    \footnotesize
    \caption{Result for supervised entity matching dataset}
    \begin{tabular}{|c|c|c|c|c|} \hline
        Dataset & Matcher & Seeds & Hit@1 & Hit@10  \\ \hline
        \multirow{9}{5em}{DW-NB} & \multirow{2}{3em}{AttrE} & 10\% & 0.88\textsuperscript{a} & 0.96\textsuperscript{a} \\
        & & 50\% & 0.88\textsuperscript{a} & 0.96\textsuperscript{a} \\ \cline{2-5}
        & \multirow{2}{3em}{NMN} & 10\% & 0.51\textsuperscript{a} & 0.60\textsuperscript{a} \\
        & & 50\% & 0.89\textsuperscript{a} & 0.95\textsuperscript{a} \\ \cline{2-5}
        & AutoAlign & 0\% & 0.89\textsuperscript{b} & \textbf{0.97}\textsuperscript{b} \\ \cline{2-5}
        & LogMap & 0\% & n/a & n/a \\ \cline{2-5}
        & BaselineAltLabel & 0\% & 0.06 & 0.06 \\ \cline{2-5}
        & PARIS & 0\% & \textbf{0.92} & {0.96} \\ \cline{2-5}
        & Our Method & 0\% & {0.91} & \textbf{0.97} \\ \hline
        
        \multirow{9}{5em}{DY-NB} & \multirow{2}{5em}{AttrE} & 10\% & 0.90\textsuperscript{a} & 0.94\textsuperscript{a} \\
        & & 50\% & 0.90\textsuperscript{a} & 0.94\textsuperscript{a} \\ \cline{2-5}
        & \multirow{2}{5em}{NMN} & 10\% & 0.56 & 0.65 \\
        & & 50\% & 0.91 & 0.95 \\ \cline{2-5}
        & AutoAlign & 0\% & 0.91\textsuperscript{b} & 0.96\textsuperscript{b} \\ \cline{2-5}
        & LogMap & 0\% & n/a & n/a \\ \cline{2-5}
        & BaselineAltLabel & 0\% & 0.07 & 0.07 \\ \cline{2-5}
        & PARIS & 0\% & {0.93} & 0.93 \\ \cline{2-5}
        & Our Method & 0\% & \textbf{0.95} & \textbf{0.97} \\ \hline
        
        \multirow{7}{5em}{DW-300k} & AttrE & 30\% & 0.71\textsuperscript{a} & 0.75\textsuperscript{a} \\ \cline{2-5}
        & NMN & 30\% & 0.71\textsuperscript{a} & 0.73\textsuperscript{a} \\ \cline{2-5}
        & AutoAlign & 0\% & n/a & n/a \\ \cline{2-5}
        & LogMap & 0\% & n/a & n/a \\ \cline{2-5}
        & BaselineAltLabel & 0\% & {0.90} & {0.90} \\ \cline{2-5}
        & PARIS & 0\% & 0.91 & 0.91 \\ \cline{2-5}
        & Our Method & 0\% & \textbf{0.94} & \textbf{0.94} \\ \hline
        
        \multirow{7}{5em}{DW-600k} & AttrE & 30\% & 0.61\textsuperscript{a} & 0.61\textsuperscript{a} \\ \cline{2-5}
        & NMN & 30\% & n/a\textsuperscript{a} & n/a\textsuperscript{a} \\ \cline{2-5}
        & AutoAlign & 0\% & n/a & n/a \\ \cline{2-5}
        & LogMap & 0\% & n/a & n/a \\ \cline{2-5}
        & BaselineAltLabel & 0\% & {0.90} & {0.90} \\ \cline{2-5}
        & PARIS & 0\% & 0.91 & 0.91 \\ \cline{2-5}
        & Our Method & 0\% & \textbf{0.93} & \textbf{0.94} \\ \hline
        \multicolumn{5}{l}{Note: \textsuperscript{a} Data from \cite{zhang2022benchmark}
        \textsuperscript{b} Data From \cite{zhang2023autoalign}}
    \end{tabular}
    
    \label{tab:result_entity_matching_supervised}
\end{table}
}

In this research, we aimed to evaluate whether incorporating triple matching could enhance performance in the entity matching task. To investigate this, we compared the results of label matching alone with the results of combining label and triple matching. Table \ref{tab:ablation} presents each method's observed values, their differences, and how many iterations the triple matching required. In most cases, triple matching either improved performance or maintained the same results, except for one instance. Notably, when label matching accuracy was already high (above 0.9), the impact of triple matching was minimal. However, in cases where label matching performance was lower, triple matching significantly boosted the results. The improvement is particularly noticeable in the DW-NB and DY-NB datasets, where the labels were intentionally manipulated to match entity pairs with different labels. For the dataset between a community-based KG and general KG (MAL-DBpedia and SWW-DBpedia), we can observe a drop in the performance for hit@1 and an improvement for hit@10. Our method can correctly find the entity with a higher possibility of match but suffers from finding the correct one. The triple matching step iterated over two to four times until stopping the process. Therefore, the additional steps require only a few iterations to increase the matching performance.

\begin{table}[]
    \centering
    \begin{tabular}{|c|c|c|c|c|c|} \hline
         Dataset & Metric & Label Matching & W/ Triple Matching & Difference & Iterations \\ \hline
         \multirow{2}{4em}{SWW-SWG} & Hit@1 & 0.77 & \textbf{0.81} & 0.04 & \multirow{2}{1em}{4} \\ \cline{2-5}
         & Hit@10 & 0.77 &  \textbf{0.83} & 0.06 & \\ \hline
         \multirow{2}{4em}{SWW-TOR} & Hit@1 & 0.93 & \textbf{0.95} & 0.02 & \multirow{2}{1em}{4} \\ \cline{2-5}
         & Hit@10 & 0.94 & \textbf{0.96} & 0.02 & \\ \hline
         \multirow{2}{4em}{MCU-MDB} & Hit@1 & 0.69 & \textbf{0.71} & 0.02 & \multirow{2}{1em}{4} \\ \cline{2-5}
         & Hit@10 & 0.72 & \textbf{0.80} & 0.08 &  \\ \hline
         \multirow{2}{4em}{MAL-MBT} & Hit@1 & \textbf{0.92} & \textbf{0.92} & - & \multirow{2}{1em}{2} \\ \cline{2-5}
         & Hit@10 & 0.94 & \textbf{0.97} & 0.03 & \\ \hline
         \multirow{2}{4em}{MAL-STX} & Hit@1 & \textbf{0.94} & \textbf{0.94} & - & \multirow{2}{1em}{2} \\ \cline{2-5}
         & Hit@10 & \textbf{0.97} & \textbf{0.97} & - & \\ \hline
         \multirow{2}{4em}{MAL-DBpedia} & Hit@1 & \textbf{0.85} & {0.75} & -0.10 & \multirow{2}{1em}{3} \\ \cline{2-5}
         & Hit@10 & {0.88} & \textbf{0.89} & 0.01 & \\ \hline
         \multirow{2}{4em}{SWW-DBpedia} & Hit@1 & \textbf{0.78} & {0.72} & -0.06 & \multirow{2}{1em}{2} \\ \cline{2-5}
         & Hit@10 & {0.81} & \textbf{0.83} & 0.02 & \\ \hline
         \multirow{2}{4em}{DW-NB} & Hit@1 & 0.60 & \textbf{0.91} & 0.31 & \multirow{2}{1em}{3}\\ \cline{2-5}
         & Hit@10 & 0.60 & \textbf{0.97} & 0.37 & \\ \hline
         \multirow{2}{4em}{DY-NB} & Hit@1 & 0.69 & \textbf{0.95} & 0.26 & \multirow{2}{1em}{2} \\ \cline{2-5}
         & Hit@10 & 0.69 & \textbf{0.97} & 0.28 & \\ \hline
         \multirow{2}{4em}{300k} & Hit@1 & \textbf{0.94} & \textbf{0.94} & - & \multirow{2}{1em}{2}\\ \cline{2-5}
         & Hit@10 & \textbf{0.94} & \textbf{0.94} & - & \\ \hline
         \multirow{2}{4em}{600k} & Hit@1 & \textbf{0.94} & {0.93} & -0.01 & \multirow{2}{1em}{2} \\ \cline{2-5}
         & Hit@10 & \textbf{0.94} & \textbf{0.94} & - & \\ \hline
    \end{tabular}
    \caption{Comparing the performance of FTM using only Label Matching and with Triple Matching}
    \label{tab:ablation}
\end{table}

\subsubsection{Triple matching} \label{subsect:results_triple_matching}

Table \ref{tab:result_triple_matching} presents the results of the triple-matching approach using the optimal threshold for each scenario. The findings show that compatible triples achieve high precision and recall across all cases. Divergent triples also demonstrate strong precision and recall; however, the SWW-SWG and MAL-MBT scenarios exhibit lower recall than others. Additionally, the results indicate that the optimal thresholds for compatible triples are notably higher than those for divergent triples.


\begin{table}[htp]
    \centering
    \footnotesize
    \caption{Result for triple matching}
        \begin{tabular}{|c|c|c|c|c|} \hline
            \multirow{2}{4em}{dataset} & \multicolumn{4}{|c|}{Compatible} \\ \cline{2-5}
            & Precision & Recall & F-measure & Best Threshold \\ \hline
            SWW-SWG & 1.00 & 0.83 & 0.91 & 0.74  \\ \hline
            SWW-TOR & 0.98 & 0.90 & 0.94 & 0.64  \\ \hline
            MAL-MBT & 0.92 & 0.86 & 0.89 & 0.38 \\ \hline
            MAL-STX & 0.86 & 0.83 & 0.84 & 0.68 \\ \hline
            MAL-DBpedia & 0.95 & 0.91 & 0.93 & 0.11 \\ \hline
            SWW-DBpedia & 0.98 & 0.66 & 0.79 & 0.48 \\ \hline
            \hline
            \multirow{2}{4em}{dataset} & \multicolumn{4}{|c|}{Divergent} \\ \cline{2-5}
            & Precision & Recall & F-measure & Best Threshold \\ \hline
            SWW-SWG & 0.75 & 0.80 & 0.77 & 0.45 \\ \hline
            SWW-TOR & 0.94 & 0.92 & 0.93 & 0.21 \\ \hline
            MAL-MBT & 0.98 & 0.76 & 0.85 & 0.19 \\ \hline
            MAL-STX & 0.93 & 0.84 & 0.89 & 0.14 \\ \hline
            MAL-DBpedia & 0.81 & 0.43 & 0.56 & 0.21 \\ \hline
            SWW-DBpedia & 0.47 & 0.63 & 0.54 & 0.27 \\ \hline
        \end{tabular}
    \label{tab:result_triple_matching}
\end{table}

\begin{figure}[ht]
    \centering
    \includegraphics[width=0.5\linewidth]{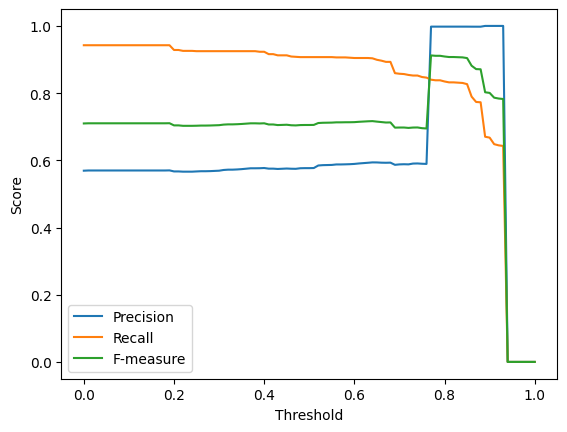}
    \caption{Triple matching threshold evolution}
    \label{fig:triple_matching}
    \Description[Triple matching threshold evolution]{}
\end{figure}

Figure \ref{fig:triple_matching} shows the threshold evolution for the SWW-TOR dataset. This dataset has stable precision, recall, and f-measure with an increase in precision with a higher threshold. In general, the datasets have similar characteristics diverging in the impact of f-measure.

    

The tables in \ref{tab:equivalent_confusion_matrix} present the confusion matrix for compatible and divergent triples using SWW-TOR. We observe a higher number of false negatives than false positives for compatible triples, mainly due to entity pairs that are not accurately mapped. In the case of divergent triples, there are fewer false positives and negatives than compatible triples. This reduction is attributed to strings with relative similarity, often resulting in a moderate score for compatibility and divergence probabilities.

\begin{table}[htp]
    \centering
    \footnotesize
    \caption{Confusion matrix for triple matching task using SWW-TOR dataset}
        \begin{tabular}{|c|c|c|c|} \hline
            Compatible& \multicolumn{3}{|c|}{Actual Values} \\ \hline
            \multirow{3}{4em}{Predicted Values} & & True & False \\ \cline{2-4}
            & True & 1021 & 16 \\ \cline{2-4}
            & False & 110 & 873 \\ \hline \hline

            Divergent & \multicolumn{3}{|c|}{Actual Values} \\ \hline
            \multirow{3}{4em}{Predicted Values} & & True & False \\ \cline{2-4}
            & True & 817 & 51 \\ \cline{2-4}
            & False & 72 & 1080 \\ \hline
        \end{tabular}
    \label{tab:equivalent_confusion_matrix}
\end{table}
\section{Discussion} \label{section:discussion}

Our approach (FTM) has comparable results and, in most cases, exceeds the current entity matchers. The dataset for unsupervised entity matching with comparable sizes (table \ref{tab:result_entity_matching}) shows that FTM has good results except for MCU-MDB. FTM depends on the information in the triples and requires the predicates to be used similarly between the KGs. The MCU-MDB does not contain matching triples for mapped entities and predicates, as observed in the triple matches in the table \ref{tab:gsStatistics}. The matching between entities can be obtained using triples with unrelated predicates but the same object value. The other matchers, such as PARIS, can explore this type of triples, maintaining a higher hit rate.

Our examination, utilizing a dataset matching real-size KGs with varying sizes and domains, demonstrates that even traditional methods face challenges in handling large KGs. However, our methodology effectively manages these KGs by leveraging the SPARQL endpoint and utilizing already matched entities and attributes exclusively.

The disparity in size between KGs also significantly influences the outcomes. For instance, employing only label matching may suffice for KG pairs with similar sizes and domains, as evidenced by the high results of the BaselineAltLabel method in all cases (Table \ref{tab:result_entity_matching}) and our comparison between the label matching and triple matching (Table \ref{tab:ablation}). However, when dealing with KGs of different sizes and domains (Table \ref{tab:result_entity_matching_real}), the performance of the matcher experiences a notable decline. A similar pattern is observed with the PARIS method in our analysis. Although FTM maintains strong results regarding hit@10, it is essential to note the decrease in hit@1 for Memory Alpha and DBpedia, indicating a potential area for improvement.

Our method improved precision and recall compared to PARIS and achieved a higher hit rate than other approaches (Table \ref{tab:result_entity_matching_real}). However, it exhibited lower performance when balancing precision and recall. This trade-off can be attributed to the design of our method, which emphasizes maximizing the retrieval of potential matches rather than strictly filtering them. In other words, FTM prioritizes coverage and inclusivity, making it particularly suitable for scenarios where the primary objective is to retrieve as many correct matches as possible, even at the cost of introducing some false positives. This characteristic is especially valuable in contexts where a broad set of candidates can subsequently be refined through human validation or downstream processing.

Overall, FTM demonstrates strong performance in recall and candidate retrieval, making it especially effective for recall-oriented entity matching tasks, where ensuring the inclusion of correct matches is more important than maximizing precision.

Considering runtime performance, PARIS demonstrated the fastest execution for the Memory Alpha and DBpedia datasets (Table \ref{tab:result_entity_matching_real}). However, it failed to complete the matching task for Star Wars Wiki and DBpedia within a month. The strong performance on Memory Alpha and DBpedia can be attributed to PARIS’s strategy of loading the entire KG into main memory and using a hash table for fast access. While this approach is efficient for smaller datasets, it is memory-intensive. In our experiments, PARIS required over 700 GB of main memory to process DBpedia. Although the dataset could be processed under these conditions, the performance was inferior to that of other methods. For Star Wars Wiki and DBpedia, PARIS struggled to explore all matching possibilities and could not complete the task within a reasonable time frame.

BaselineAltLabel could potentially achieve faster runtimes by also loading the KG into memory; however, in our evaluation, we chose to dynamically load data from a SPARQL endpoint. Similarly, our proposed approach retrieves data dynamically from the endpoint instead of loading the entire DBpedia KG into memory. While this results in slower runtimes for smaller datasets, it ensures a more consistent and scalable performance across datasets of varying sizes, making it more appropriate for larger dataset. Moreover, our method incorporates an early stopping mechanism to reduce the number of iterations and improve efficiency.

We compared the results for supervised methods, such as AttrE and NMN, with those of statistical methods like LogMap, PARIS, and FTM in the table \ref{tab:result_entity_matching_supervised}. Comparing statistical and supervised methods, it is evident that statistical methods consistently yield higher results than supervised methods across all scenarios, particularly in hit@1 scenarios. AttrE performs similarly to unsupervised methods across two datasets, while NMN requires substantial training data to achieve similar outcomes. The difference between supervised and statistical methods is evident, especially for scalability datasets (DW-300k and DW-600k). The accuracy of supervised methods degrades when the number of entities increases, as observed by AttrE results. NMN has a similar result to AttrE for 300k but could not process 600k due to memory requirements. All statistical methods have stable results across all supervised method datasets. 
 
These findings underscore the superior robustness of statistical methods over their supervised counterparts, highlighting the potential for further enhancement, especially in GCN methods. Notably, straightforward approaches like exact matches, as seen in BaselineAltLabel, can attain competitive results with the current dataset. It shows a need for datasets that rigorously assess matching capabilities across various scales and domains. On the other hand, matching approaches that use supervised methods should be tested, comparing not only with other supervised methods but with statistical methods.

For entity matching, setting a threshold is not required when using the hit rate as the evaluation metric. Our primary goal in this context is to achieve higher recall, as emphasized by the use of the hit rate. However, a threshold becomes essential when optimizing the F-measure. The optimal threshold is one that improves precision while minimizing the negative impact on recall. As shown in Tables \ref{tab:result_entity_matching} and \ref{tab:result_entity_matching_real}, our method achieves its best F-measure with thresholds in the range of 0.90 to 0.94. These thresholds were determined by exhaustively evaluating all possible values to identify the ones that maximize the F-measure, and the selected thresholds are presented in the tables. Although higher thresholds tend to reduce recall, they result in competitive levels of precision compared to the baselines.

The triple matching approach demonstrates high precision, recall, and F-measure, particularly for compatible triples as outline in the table \ref{tab:result_triple_matching}. As entity matching and triple matching are interdependent, the quality of triple matching significantly affects the quality of entity matching. For instance, in the MAL-MBT dataset, over 70\% of false negatives involving divergent triples were due to missing entity pairs.

Challenges arise when KGs use properties in varying formats. For instance, SWW and DBpedia demonstrate lower recall for compatible and divergent triple matching due to differences in predicate representation. In SWW, the predicate \textit{nationality} links to a demonym as a string (e.g., American). In contrast, DBpedia links to a country as an entity (e.g., dbp:United\_States), reducing the similarity between the two. Furthermore, websites are often processed as entities, adding complexity to data integration. While entity matching can explore different properties to establish correspondences, object matches frequently contain limited information, necessitating external resources to enhance these matches.

Our dataset includes instances with limited examples. While SWW-SWG, MAL-DBpedia, and SWW-DBpedia exhibit sufficient entity matches, only a few properties were successfully matched and utilized. Consequently, these cases contain fewer than a hundred examples and perform less than other scenarios. This underperformance can be attributed to factors such as unmapped entity pairs, highlighting the challenges in evaluating these cases effectively.

{
\begin{table*}[htp]
    \centering
    \scriptsize
    \begin{tabular}{|c|p{.2\textwidth}p{.2\textwidth}p{.15\textwidth}|c|c|} \hline
        Number & Subject & Predicate & Objects & Expected result & Generated result \\ \hline \hline
        \multirow{2}{2em}{1} & \url{memory-alpha:resource/Behind_Enemy_Lines} & \url{memory-alpha:property/pages} & 288 & \multirow{2}{4em}{Compatible} & \multirow{2}{4em}{Compatible} \\
        & \url{memory-beta:resource/Behind_Enemy_Lines} & \url{memory-beta:property/pages} & 269 & & \\ \hline \hline
        
        \multirow{2}{2em}{2} &  \url{memory-alpha:resource/Stellar_Cartography:_The_Starfleet_Reference_Library} & \url{memory-alpha:property/pages} & 48 & \multirow{2}{4em}{Divergent} &\multirow{2}{4em}{Compatible} \\
        & \url{memory-beta:resource/Stellar_Cartography:_The_Starfleet_Reference_Library} & \url{memory-beta:property/pages} & 4810 & & \\ \hline \hline
        
        \multirow{2}{2em}{3} & \url{memory-alpha:resource/Best_of_Captain_Kirk} & \url{memory-alpha:property/prev} & \url{memory-alpha:resource/Best_of_Star_Trek:_Deep_Space_Nine} & \multirow{2}{4em}{Compatible} & \multirow{2}{4em}{Compatible} \\
        & \url{memory-beta:resource/Best_of_Captain_Kirk} & \url{memory-beta:property/before} & \url{memory-beta:property/Best_of_DS9} & & \\ \hline \hline
        
        \multirow{2}{2em}{4} & \url{memory-alpha:resource/Simon_Tarses} & \url{memory-alpha:property/species} & \url{memory-alpha:resource/Human} & \multirow{2}{4em}{Divergent} & \multirow{2}{4em}{Divergent} \\
        & \url{memory-beta:resource/Simon_Tarses} & \url{memory-beta:property/species} & 3 & & \\ \hline \hline

        \multirow{2}{2em}{5} & \url{memory-alpha:resource/USS_Aries} & \url{memory-alpha:property/registry} & \url{memory-alpha:resource/NCC} & \multirow{2}{4em}{Divergent} & \multirow{2}{4em}{Compatible} \\
        & \url{memory-beta:resource/USS_Aries_(NCC-45167)} & \url{memory-beta:property/registry} & NCC-45167 & & \\ \hline \hline

        \multirow{2}{2em}{6} & \url{memory-alpha:resource/Bruce_Maddox} & \url{memory-alpha:property/rank} & \url{memory-alpha:resource/Commander} & \multirow{2}{4em}{Divergent} & \multirow{2}{4em}{Compatible} \\
        & \url{memory-beta:resource/Bruce_Maddox} & \url{memory-beta:property/rank} & \url{memory-beta:resource/Captain} & & \\ \hline 

        \multirow{2}{2em}{7} & \url{memory-alpha:resource/Khan_Noonien_Singh_(alternate_reality)} & \url{memory-alpha:property/species} & \url{memory-alpha:resource/Human} & \multirow{2}{4em}{Compatible} & \multirow{2}{4em}{Compatible} \\
        & \url{memory-beta:resource/Khan_Noonien_Singh_(alternate_reality)} & \url{memory-beta:property/species} & \url{memory-beta:resource/Human} & & \\ \hline \hline

        \multirow{2}{2em}{8} & \url{memory-alpha:resource/Khan_Noonien_Singh_(alternate_reality)} & \url{memory-alpha:property/species} & \url{memory-alpha:resource/Augment} & \multirow{2}{4em}{Divergent} & \multirow{2}{4em}{Divergent} \\
        & \url{memory-beta:resource/Khan_Noonien_Singh_(alternate_reality)} & \url{memory-beta:property/species} & \url{memory-beta:resource/Human} & & \\ \hline
    \end{tabular}
    \caption{Example of mapping between triples}
    
    \label{tab:example_numbers}
\end{table*}
}

The main objective of our proposed method is to calculate the probability of two triples being compatible or divergent because we consider that KGs can be noisy. However, to test FTM, we must classify the triple mappings as compatible or divergent, so there is a subjective element to deciding if two triples are compatible. The table \ref{tab:example_numbers} has examples of matching triples. Example 1 shows a case where the difference between the numbers is small, so they are compatible. However, example 2 is a case in which the difference is larger, so they are considered divergent and must be verified.

Another case of divergent information is when the type does not match. For example, we expect cases like example \ref{tab:example_numbers}.3, which has matching entities linked to the species. However, we have cases like \ref{tab:example_numbers}.4, which has an entity related to species in the first triple but a number linked to species in the second. This mapping can be classified as divergent and must be verified.

The current approach has issues with false positives when the triple mapping has similar strings or the entities are related. The triples in the example \ref{tab:example_numbers}.5 are classified as compatibles because the label from entity \textit{\url{memory-alpha:resource/NCC}} is \textit{NCC} and the similarity with \textit{NCC-45167} is 0.9. However, the entity refers to the registry number prefix, not the registry itself, so they do not represent the same information. We can not define which one is correct because the difference can be caused by differences in how each community uses the predicate. Still, we can not classify them as compatibles.

Another case of false positive is triples in the example \ref{tab:example_numbers}.6. The entities \textit{\url{memory-alpha:resource/Commander}} and \textit{\url{memory-beta:resource/Captain}} are related and have similarity 0.66 in the entity mapping, but they refer to different rank, so they can not be classified as compatibles.

Our current approach uses functionality to select predicates that usually have only one triple related to the same entity and predicate, but we have some exceptions. For example, each entity usually has only one species, so the functionality is 0.96. However, we can observe that species of \textit{Khan Noonien Singh} is both \textit{Human} and \textit{Augment} in Memory Alpha KG. In the Memory Beta KG, we only have triple referring to \textit{Human}, so the example \ref{tab:example_numbers}.7 is compatible, and the example \ref{tab:example_numbers}.8 is divergent, but both information is correct. This issue can be more problematic when both KGs have this kind of exception because we use cross-products to match triples with the same entity and predicate. This is one of the reasons why we do not infer which one is the correct one. We can extend the current approach to solve these local exceptions, detect outliers, or use multiple KGs with the same triples, but these are not tested in the current state.

Triple matching relies on the use of thresholds to optimize the F-measure. Figure \ref{fig:triple_matching} illustrates the typical behavior of precision, recall, and F-measure as the threshold varies. Lower threshold values tend to yield higher recall but lower precision. As the threshold increases, precision and F-measure initially improve, until reaching a point beyond which performance declines.

To select the most appropriate threshold, we perform an exhaustive search over a range of possible values and select the one that yields the highest F-measure on a validation set. This method is straightforward and ensures that the selected threshold is optimal with respect to F-measure. While the performance is somewhat sensitive to the choice of threshold, small variations around the optimal value typically result in minor changes in performance. Moreover, we observe that thresholds corresponding to compatible triples tend to be higher, whereas those for divergent triples are lower, which can also guide threshold tuning in practical applications.

The results of the triple matching process demonstrate its effectiveness in identifying and distinguishing compatible and divergent triples between KGs. This information enhances entity matching and highlights differences in how triples are utilized across KGs. These differences often originated from variations in the contextual representation within the KGs or disparities in the quality of the extracted information. Identifying such differences is particularly valuable for integrating heterogeneous KGs, especially when constructed from diverse sources and backgrounds, as is frequently the case with community-driven KGs. Furthermore, we provided examples of mappings derived from the triple matching, illustrating the method’s success and revealing areas for potential improvement. These findings underscore the significance of triple matching as a tool for advancing effective KG integration and addressing challenges in diverse and complex data environments.

\section{Conclusion} \label{section:conclusion}

This study introduces an innovative method for discovering mappings between triples in two heterogeneous Knowledge Graphs (KGs). Our approach, consisting of label matching and triple matching stages, creates the mapping triples that support the entity mapping. 
Validation of our method on different datasets shows that Full Triple Matcher (FTM) is consistently comparable to outperforming the results for existing state-of-the-art entity matching methods. We also show that statistical entity-matching methods can have state-of-the-art results and beat the current supervised entity-matching methods, especially in the larger dataset that more closely replicates real-world scenarios.

Our proposed technique demonstrates high precision in accurately matching and categorizing triples, distinguishing between compatible and divergent information. Through illustrative examples, we showcase successful mappings generated by FTM for triples with both compatible and divergent information. This research represents a crucial step toward integrating diverse KGs by generating a set of consistent triples.

Furthermore, our proposed methods enhance KG accuracy by cross-referencing information across different KGs, thereby identifying and addressing potential conflicts. This approach holds promise for applications in question answering, where it can expand KG coverage and provide alternative answers tailored to specific accuracy requirements or contextual nuances.

For future work, we want to explore different approaches to search for similar labels in the entity-matching method and solve the current limitation of only using 1-1 entity-predicate triples. We plan to expand our current approach to explore N-N KG mappings to automatically identify and correct wrong values or propose changes for human intervention. Additionally, we aim to explore entity matching for multimodal knowledge graphs, considering how to leverage data types beyond textual types, such as images and videos, for improved alignment. Furthermore, we want to explore how to integrate the mapped triples from different KGs in a concise and easy-to-use endpoint.

\printbibliography

@inproceedings{hertling2018,
  title={Dbkwik: A consolidated knowledge graph from thousands of wikis},
  author={Hertling, Sven and Paulheim, Heiko},
  booktitle={2018 IEEE International Conference on Big Knowledge (ICBK)},
  pages={17--24},
  year={2018},
  organization={IEEE}
}

@misc{bachmann2023,
  author        = {Bachmann, Max},
  year          = {2023},
  title         = {Rapid fuzzy string matching in Python and C++ using the Levenshtein Distance},
  note          = {Accessed August 11, 2023, https://github.com/maxbachmann/RapidFuzz}
}

@article{noia2016,
author = {Noia, Tommaso Di and Ostuni, Vito Claudio and Tomeo, Paolo and Sciascio, Eugenio Di},
title = {SPrank: Semantic Path-Based Ranking for Top-N Recommendations Using Linked Open Data},
year = {2016},
issue_date = {January 2017},
publisher = {Association for Computing Machinery},
address = {New York, NY, USA},
volume = {8},
number = {1},
issn = {2157-6904},
url = {https://doi.org/10.1145/2899005},
doi = {10.1145/2899005},
journal = {ACM Trans. Intell. Syst. Technol.},
month = {9},
articleno = {9},
numpages = {34}
}

@article{leone2022,
author = {Leone, Manuel and Huber, Stefano and Arora, Akhil and Garc\'{\i}a-Dur\'{a}n, Alberto and West, Robert},
title = {A Critical Re-Evaluation of Neural Methods for Entity Alignment},
year = {2022},
issue_date = {April 2022},
publisher = {VLDB Endowment},
volume = {15},
number = {8},
issn = {2150-8097},
url = {https://doi.org/10.14778/3529337.3529355},
doi = {10.14778/3529337.3529355},
journal = {Proc. VLDB Endow.},
month = {4},
pages = {1712–1725},
numpages = {14}
}

@InProceedings{Liu2017,
author="Liu, Shuangyan
and d'Aquin, Mathieu
and Motta, Enrico",
editor="Gracia, Jorge
and Bond, Francis
and McCrae, John P.
and Buitelaar, Paul
and Chiarcos, Christian
and Hellmann, Sebastian",
title="Measuring Accuracy of Triples in Knowledge Graphs",
booktitle="Language, Data, and Knowledge",
year="2017",
publisher="Springer International Publishing",
address="Cham",
pages="343--357",
isbn="978-3-319-59888-8"
}

@inproceedings{steven2004,
    title = "{NLTK}: The Natural Language Toolkit",
    author = "Bird, Steven  and
      Loper, Edward",
    booktitle = "Proceedings of the {ACL} Interactive Poster and Demonstration Sessions",
    month = jul,
    year = "2004",
    address = "Barcelona, Spain",
    publisher = "Association for Computational Linguistics",
    url = "https://aclanthology.org/P04-3031",
    pages = "214--217",
}

@article{suchanek2011,
author = {Suchanek, Fabian M. and Abiteboul, Serge and Senellart, Pierre},
title = {PARIS: Probabilistic Alignment of Relations, Instances, and Schema},
year = {2011},
issue_date = {November 2011},
publisher = {VLDB Endowment},
volume = {5},
number = {3},
issn = {2150-8097},
url = {https://doi.org/10.14778/2078331.2078332},
doi = {10.14778/2078331.2078332},
journal = {Proc. VLDB Endow.},
month = {11},
pages = {157–168},
numpages = {12}
}

@inproceedings{hertling2020,
  title={The knowledge graph track at OAEI: Gold standards, baselines, and the golden hammer bias},
  author={Hertling, Sven and Paulheim, Heiko},
  booktitle={The Semantic Web: 17th International Conference, ESWC 2020, Heraklion, Crete, Greece, May 31--June 4, 2020, Proceedings 17},
  pages={343--359},
  year={2020},
  organization={Springer}
}

@article{hertling2020dbkwik,
  title={Dbkwik: extracting and integrating knowledge from thousands of wikis},
  author={Hertling, Sven and Paulheim, Heiko},
  journal={Knowledge and Information Systems},
  volume={62},
  number={6},
  pages={2169--2190},
  year={2020},
  publisher={Springer}
}

@inproceedings{pour2023a,
  booktitle={Ontology Matching},
  edition = {},
  number = {},
  journal = {},
  pages = {97-139},
  publisher = {CEUR Workshop Proceedings},
  school = {},
  title = {Results of the Ontology Alignment Evaluation Initiative 2023},
  volume = {},
  author = {Mina Abd Nikooie Pour and Alsayed Algergawy and Patrice Buche and Leyla J. Castro and Jiaoyan Chen and Adrien Coulet and Julien Cufi and Hang Dong and Omaima Fallatah and Daniel Faria and Irini Fundulaki and Sven Hertling and Yuan He and Ian Horrocks and Martin Huschka and Liliana Ibanescu and Sarika Jain and Ernesto Jiménez-Ruiz and Naouel Karam and Patrick Lambrix and Huanyu Li and Ying Li and Pierre Monnin and Engy Nasr and Heiko Paulheim and Catia Pesquita and Tzanina Saveta and Pavel Shvaiko and Guilherme Sousa and Cássia Trojahn and Jana Vatascinova and Mingfang Wu and Beyza Yaman and Ondřej Zamazal and Lu Zhou},
  editor = {},
  year = {2023},
  organizer = {18th International Workshop on Ontology Matching (OM 2023)},
  series = {CEUR Workshop Proceedings}
}

@inproceedings{koudas2009metric,
  title={Metric functional dependencies},
  author={Koudas, Nick and Saha, Avishek and Srivastava, Divesh and Venkatasubramanian, Suresh},
  booktitle={2009 IEEE 25th International Conference on Data Engineering},
  pages={1275--1278},
  year={2009},
  organization={IEEE}
}

@INPROCEEDINGS{munne2023,
  author={Munne, Rumana Ferdous and Ichise, Ryutaro},
  booktitle={2023 IEEE 17th International Conference on Semantic Computing (ICSC)}, 
  title={Attribute Enhancement using Aligned Entities between Knowledge Graphs}, 
  year={2023},
  volume={},
  number={},
  pages={191-198},
  doi={10.1109/ICSC56153.2023.00038}}

@article{hertling2022gollum,
  title={Gollum: A gold standard for large scale multi source knowledge graph matching},
  author={Hertling, Sven and Paulheim, Heiko},
  journal={arXiv preprint arXiv:2209.07479},
  year={2022}
}

@article{jiang2023rethinking,
  title={Rethinking GNN-based Entity Alignment on Heterogeneous Knowledge Graphs: New Datasets and A New Method},
  author={Jiang, Xuhui and Xu, Chengjin and Shen, Yinghan and Su, Fenglong and Wang, Yuanzhuo and Sun, Fei and Li, Zixuan and Shen, Huawei},
  journal={arXiv preprint arXiv:2304.03468},
  year={2023}
}

@article{zhang2022benchmark,
  title={A benchmark and comprehensive survey on knowledge graph entity alignment via representation learning},
  author={Zhang, Rui and Trisedya, Bayu Distiawan and Li, Miao and Jiang, Yong and Qi, Jianzhong},
  journal={The VLDB Journal},
  volume={31},
  number={5},
  pages={1143--1168},
  year={2022},
  publisher={Springer}
}

@inproceedings{trisedya2019entity,
  title={Entity alignment between knowledge graphs using attribute embeddings},
  author={Trisedya, Bayu Distiawan and Qi, Jianzhong and Zhang, Rui},
  booktitle={Proceedings of the AAAI conference on artificial intelligence},
  volume={33},
  number={01},
  pages={297--304},
  year={2019}
}

@inproceedings{wu-2020-neighborhood,
    title = "Neighborhood Matching Network for Entity Alignment",
    author = "Wu, Yuting  and
      Liu, Xiao  and
      Feng, Yansong  and
      Wang, Zheng  and
      Zhao, Dongyan",
    editor = "Jurafsky, Dan  and
      Chai, Joyce  and
      Schluter, Natalie  and
      Tetreault, Joel",
    booktitle = "Proceedings of the 58th Annual Meeting of the Association for Computational Linguistics",
    month = jul,
    year = "2020",
    address = "Online",
    publisher = "Association for Computational Linguistics",
    url = "https://aclanthology.org/2020.acl-main.578",
    doi = "10.18653/v1/2020.acl-main.578",
    pages = "6477--6487",
}

@article{bordes2013translating,
  title={Translating embeddings for modeling multi-relational data},
  author={Bordes, Antoine and Usunier, Nicolas and Garcia-Duran, Alberto and Weston, Jason and Yakhnenko, Oksana},
  journal={Advances in neural information processing systems},
  volume={26},
  year={2013}
}

@inproceedings{jimenez2011logmap,
  title={Logmap: Logic-based and scalable ontology matching},
  author={Jim{\'e}nez-Ruiz, Ernesto and Cuenca Grau, Bernardo},
  booktitle={The Semantic Web--ISWC 2011: 10th International Semantic Web Conference, Bonn, Germany, October 23-27, 2011, Proceedings, Part I 10},
  pages={273--288},
  year={2011},
  organization={Springer}
}

@article{devlin2018bert,
  author    = {Jacob Devlin and
               Ming{-}Wei Chang and
               Kenton Lee and
               Kristina Toutanova},
  title     = {{BERT:} Pre-training of Deep Bidirectional Transformers for Language
               Understanding},
  journal   = {CoRR},
  volume    = {abs/1810.04805},
  year      = {2018},
  url       = {http://arxiv.org/abs/1810.04805},
  archivePrefix = {arXiv},
  eprint    = {1810.04805},
  bibsource = {dblp computer science bibliography, https://dblp.org}
}

@article{zhang2023autoalign,
  title={Autoalign: fully automatic and effective knowledge graph alignment enabled by large language models},
  author={Zhang, Rui and Su, Yixin and Trisedya, Bayu Distiawan and Zhao, Xiaoyan and Yang, Min and Cheng, Hong and Qi, Jianzhong},
  journal={IEEE Transactions on Knowledge and Data Engineering},
  year={2023},
  publisher={IEEE}
}

@inproceedings{huang2024representation,
  title={Representation Learning for Entity Alignment in Knowledge Graph: A Design Space Exploration},
  author={Huang, Peng and Zhang, Meihui and Zhong, Ziyue and Chai, Chengliang and Fan, Ju},
  booktitle={2024 IEEE 40th International Conference on Data Engineering (ICDE)},
  pages={3462--3475},
  year={2024},
  organization={IEEE}
}

@article{Huang2023DAAKG,
author = {Huang, Jiacheng and Sun, Zequn and Chen, Qijin and Xu, Xiaozhou and Ren, Weijun and Hu, Wei},
title = {Deep Active Alignment of Knowledge Graph Entities and Schemata},
year = {2023},
issue_date = {June 2023},
publisher = {Association for Computing Machinery},
address = {New York, NY, USA},
volume = {1},
number = {2},
url = {https://doi.org/10.1145/3589304},
doi = {10.1145/3589304},
journal = {Proc. ACM Manag. Data},
month = jun,
articleno = {159},
numpages = {26}
}

@article{OpenEA,
  author    = {Zequn Sun and
               Qingheng Zhang and
               Wei Hu and
               Chengming Wang and
               Muhao Chen and
               Farahnaz Akrami and
               Chengkai Li},
  title     = {A Benchmarking Study of Embedding-based Entity Alignment for Knowledge Graphs},
  journal   = {Proceedings of the VLDB Endowment},
  volume    = {13},
  number    = {11},
  pages     = {2326--2340},
  year      = {2020},
  url       = {http://www.vldb.org/pvldb/vol13/p2326-sun.pdf}
}

@InProceedings{pmlr-v162-guo22i,
  title = 	 {Understanding and Improving Knowledge Graph Embedding for Entity Alignment},
  author =       {Guo, Lingbing and Zhang, Qiang and Sun, Zequn and Chen, Mingyang and Hu, Wei and Chen, Huajun},
  booktitle = 	 {Proceedings of the 39th International Conference on Machine Learning},
  pages = 	 {8145--8156},
  year = 	 {2022},
  editor = 	 {Chaudhuri, Kamalika and Jegelka, Stefanie and Song, Le and Szepesvari, Csaba and Niu, Gang and Sabato, Sivan},
  volume = 	 {162},
  series = 	 {Proceedings of Machine Learning Research},
  month = 	 {7},
  publisher =    {PMLR},
  url = 	 {https://proceedings.mlr.press/v162/guo22i.html}
}

@article{hogan2021knowledge,
  title={Knowledge graphs},
  author={Hogan, Aidan and Blomqvist, Eva and Cochez, Michael and d’Amato, Claudia and Melo, Gerard De and Gutierrez, Claudio and Kirrane, Sabrina and Gayo, Jos{\'e} Emilio Labra and Navigli, Roberto and Neumaier, Sebastian and others},
  journal={ACM Computing Surveys (Csur)},
  volume={54},
  number={4},
  pages={1--37},
  year={2021},
  publisher={ACM New York, NY, USA}
}

@InProceedings{GuhaContext2004,
author="Guha, Ramanathan
and McCool, Rob
and Fikes, Richard",
editor="McIlraith, Sheila A.
and Plexousakis, Dimitris
and van Harmelen, Frank",
title="Contexts for the Semantic Web",
booktitle="The Semantic Web -- ISWC 2004",
year="2004",
publisher="Springer Berlin Heidelberg",
address="Berlin, Heidelberg",
pages="32--46",
isbn="978-3-540-30475-3"
}

@incollection{shvaiko2005survey,
  title={A survey of schema-based matching approaches},
  author={Shvaiko, Pavel and Euzenat, J{\'e}r{\^o}me},
  booktitle={Journal on data semantics IV},
  pages={146--171},
  year={2005},
  publisher={Springer}
}

@article{alwan2017survey,
  title={A survey of schema matching research using database schemas and instances},
  author={Alwan, Ali A and Nordin, Azlin and Alzeber, Mogahed and Abualkishik, Abedallah Zaid},
  journal={International Journal of Advanced Computer Science and Applications},
  volume={8},
  number={10},
  year={2017},
  publisher={Science and Information (SAI) Organization Limited}
}

@inproceedings{madhavan2001generic,
  title={Generic schema matching with cupid},
  author={Madhavan, Jayant and Bernstein, Philip A and Rahm, Erhard},
  booktitle={vldb},
  volume={1},
  number={2001},
  pages={49--58},
  year={2001}
}

@article{cruz2009agreementmaker,
  title={AgreementMaker: efficient matching for large real-world schemas and ontologies},
  author={Cruz, Isabel F and Antonelli, Flavio Palandri and Stroe, Cosmin},
  journal={Proceedings of the VLDB Endowment},
  volume={2},
  number={2},
  pages={1586--1589},
  year={2009},
  publisher={VLDB Endowment}
}

@article{fang2024mdsea,
  title={MDSEA: Knowledge Graph Entity Alignment Based on Multimodal Data Supervision},
  author={Fang, Jianyong and Yan, Xuefeng},
  journal={Applied Sciences},
  volume={14},
  number={9},
  pages={3648},
  year={2024},
  publisher={MDPI}
}

@article{wang2025multi,
  title={Multi-modal Entity Alignment Based on Multidimensional Semantic Extraction},
  author={Wang, Huansha and Liu, Qinrang and Huang, Ruiyang and Zhang, Jianpeng and Liu, Hongji},
  journal={IEICE Transactions on Information and Systems},
  pages={2024EDP7173},
  year={2025},
  publisher={The Institute of Electronics, Information and Communication Engineers}
}

@article{yang2024two,
  title={Two heads are better than one: Integrating knowledge from knowledge graphs and large language models for entity alignment},
  author={Yang, Linyao and Chen, Hongyang and Wang, Xiao and Yang, Jing and Wang, Fei-Yue and Liu, Han},
  journal={arXiv preprint arXiv:2401.16960},
  year={2024}
}

@article{chen2024llm,
  title={LLM-Align: Utilizing Large Language Models for Entity Alignment in Knowledge Graphs},
  author={Chen, Xuan and Lu, Tong and Wang, Zhichun},
  journal={arXiv preprint arXiv:2412.04690},
  year={2024}
}

\end{document}